\documentclass[12pt,a4paper]{article}
\usepackage[T1]{fontenc}
\usepackage[utf8]{inputenc}
\usepackage{float}
\usepackage{url}
\usepackage{natbib}
\usepackage{authblk}
\usepackage{amsfonts}
\usepackage{amsmath}
\usepackage{amssymb}
\usepackage{graphicx}
\usepackage{caption}
\usepackage{subcaption}
\usepackage{nameref}
\usepackage{multirow}
\usepackage{booktabs}
\usepackage{adjustbox}

\let\originalleft\left
\let\originalright\right
\renewcommand{\left}{\mathopen{}\mathclose\bgroup\originalleft}
\renewcommand{\right}{\aftergroup\egroup\originalright}
\newcommand{\smallspace}{\;\;}

\newcommand{\mat}[1]{\mathbf{#1}}

\newcommand{\A}{\mathbf{A}}

\newcommand{\x}{\mathbf{x}}

\newcommand{\Y}{\mathbf{Y}}

\begin{document}
	\title{Drug response prediction by inferring pathway-response associations with Kernelized Bayesian Matrix Factorization}
	\author[1]{Muhammad Ammad-ud-din\thanks{muhammad.ammad-ud-din@aalto.fi}} 
	\author[2]{Suleiman A.Khan}
	\author[2]{Disha Malani}  
	\author[2]{Astrid Murum\"{a}gi}
	\author[2,3,4]{Olli Kallioniemi}
	\author[2,5]{Tero Aittokallio}
	\author[1]{Samuel Kaski\thanks{samuel.kaski@aalto.fi}}
	\affil[1]{Helsinki Institute for Information Technology HIIT, Department of Computer Science, Aalto University, 02150 Espoo, Finland.}
	\affil[2]{Institute for Molecular Medicine Finland FIMM, University of Helsinki, 00290 Helsinki, Finland.}
	\affil[3]{Science for Life Laboratory,}
	\affil[4]{Department of Oncology and Pathology, Karolinska Institutet, 17165 Solna, Sweden.}
	\affil[5]{Department of Mathematics and Statistics, University of Turku, 20014 Turku, Finland.}
	\maketitle
	\begin{abstract}	
		A key goal of computational personalized medicine is to systematically utilize genomic and other molecular features of samples to predict drug responses for a previously unseen sample. Such predictions are valuable for developing hypotheses for selecting therapies tailored for individual patients. This is especially valuable in oncology, where molecular and genetic heterogeneity of the cells has a major impact on the response. However, the prediction task is extremely challenging, raising the need for methods that can effectively model and predict drug responses.
		In this study, we propose a novel formulation of multi-task matrix factorization that allows selective data integration for predicting drug responses. To solve the modeling task, we extend the state-of-the-art kernelized Bayesian matrix factorization (KBMF) method with component-wise multiple kernel learning. In addition, our approach exploits the known pathway information in a novel and biologically meaningful fashion to learn the drug response associations. Our method quantitatively outperforms the state of the art on predicting drug responses in two publicly available cancer data sets as well as on a synthetic data set. In addition, we validated our model predictions with lab experiments using an in-house cancer cell line panel. We finally show the practical applicability of the proposed method by utilizing prior knowledge to infer pathway-drug response associations, opening up the opportunity for elucidating drug action mechanisms. We demonstrate that pathway-response associations can be learned by the proposed model for the well known EGFR and MEK inhibitors. 	
	\end{abstract}
	
	\section{Introduction}
	The fundamental aim of personalized medicine is to design and identify individualized therapies that maximize drug efficacy while minimizing the undesirable side effects. The efficacy, however, depends on a multitude of factors, including molecular, genetic, environmental and clinical characteristics of the samples, and much of this information remains unknown. A promising research direction is to computationally learn to predict, based on the available molecular and genetic descriptions of the samples, the responses they elicit in lab when exposed to a spectrum of drugs. The learned predictors help identifying potential drug response associations, and can predict responses for a new sample.    
	
	The development of molecular and genetic models of drug response has been made possible through several recent large scale high-throughput screening efforts that profile large panels of human cancer cell lines and drugs \citep{barretina2012cancer,Garnett2012,basu2013interactive}. Such models open up the opportunity to study the impact of molecular characteristics on the response, increasing our understanding of cancer vulnerabilities as well as making it possible to build predictive models of drug responsiveness. 
	
	Recent advances have demonstrated that molecular and genomic features have been useful in predicting the drug responses in cell lines \citep{costello2014community,jang2014systematic}. However, a key challenge underlying predictive modeling is the small sample size and very large number of genomic features. The small sample sizes offer limited statistical power leading to high uncertainty in the predictions. The inherent heterogeneity across and within different cancer types makes robust inference even harder. In the absence of technical and practical facilities to overcome these limitations, a prospective direction is to incorporate additional prior knowledge in a biologically meaningful way to facilitate the learning process.
	
	From the computational perspective, several methodologies have been used to predict drug response (a detailed discussion in Section ~\ref{sec:rel-work}). A key constituent of inferring the molecular and genetic model is the ability to effectively integrate multiple side-data views (also called as side-data sources) for prediction of the drug responses. Methods commonly referred to as multiple kernel learning MKL~\citep{gonen2011multiple} can extract the common signal from multiple side-data views, effectively yielding an increased signal-to-noise ratio in the parameter space and are currently the state of the art in drug response prediction~\citep{costello2014community}. Multi-task learning makes it possible to learn a predictive model for all of the drugs jointly (multi-task) making it possible to gather statistical evidence across multiple drugs \citep{baxter2000model}. 
	
	In this study we introduce component-wise multiple kernel learning (MKL) into the recent kernelized Bayesian matrix factorization (KBMF) method~\citep{gonen2013kernelized}. The proposed model solves the prediction task by gathering evidence from multiple side-data views, selectively, for each of the output variable group. This formulation is particularly useful in drug response prediction, for taking into account multiple side-data views. It need not assume the same views to be relevant to all drugs as earlier methods, but instead predictions can be based on different views for different groups of drugs.
	The multiple side-data views can be generated based on the prior biological knowledge; in the paper we use the pathways that are linked to the known primary targets of the drugs. By systematically utilizing this type of prior knowledge through kernelized Bayesian matrix factorization with the component-wise MKL approach, we hypothesize that pathway-drug response associations can be learned which are more informative for response prediction, and additionally are better interpretable for understanding drug action mechanisms. 
	\subsection*{Contributions} 
	In this paper, we present a novel approach for improving accuracy of predicting drug responses and elucidating the underlying pathway-drug response associations. Specifically, our contributions are two-fold: 
	\begin{enumerate}
		\item Methodologically we extend the current state-of-the-art model kernelized Bayesian matrix factorization (KBMF) with component-wise multiple kernel learning (MKL). The extension can be seen as multi-task learning by task factorization, however with selective data integration. Here the key assumption is that component-wise MKL allows the method to better use prior biological knowledge (pathways) input as multiple side-data views.
		\item We introduce a way for incorporating prior biological knowledge, in the form of pathways, for modeling pathway-drug response associations. Instead of using a single side-data view for the genomic features, we present pathway-based groups of features as multiple side-data views. Here the key assumption is that informed grouping of the features introduces additional structure and knowledge that is valuable for prediction of particular drug groups. 
	\end{enumerate}
	We first demonstrate the model's predictive abilities on a synthetic data set. We then substantiate the significantly better performance of our approach  on predicting drug responses in two large publicly available cancer data sets. In addition, we validate the \emph{in silico} predictions of our model with lab experiments on an in-house Acute Myeloid Leukemia (AML) cell line panel. Finally, we examine the inferred associations between drug responses and pathways in the larger data set, demonstrating a mechanism for elucidating drug action mechanisms.
	\section{Related Work in drug response prediction}\label{sec:rel-work}
	The computational task underlying personalized medicine is to predict drug responses on new cancer cell lines, given a set of cancer cell lines for which some measurements of drug responses are observed.
	
	A common approach is to use the mean of the observed responses as predictions for the unobserved (unseen) drug responses (used as baseline method here). Another well-known supervised approach uses the genomic and molecular features of the cell lines (as input side-data) and the observed drug responses to learn a predictive model of the drug responses~\citep{jang2014systematic}. The available molecular and genomic features range from gene expression to copy number and point mutations for the cancer cell lines, respectively~\citep{barretina2012cancer,Garnett2012}.
	
	Another widely used approach is the quantitative structure-activity relationship {\em (QSAR)} analysis which uses chemical and structural properties (often called as descriptors) of the drugs and the observed responses to learn a predictive model to infer the unobserved responses. The descriptors vary from 2D fingerprints to spatial characteristics and physiochemical features of the drugs~\citep{Perkins2003,Myint2010,Shao2012}. Recently, an advanced approach has been proposed that learns a joint predictive model of the observed drug responses by combining both the genomic features of the cell lines and descriptors of the drugs~\citep{menden2013machine,ammad2014integrative,cortes2015improved, zhang2015predicting, cichonska2015identification}.
	
	Previous studies have used linear as well as non-linear methods.
	Linear methods including multivariate linear regression, partial least squares (PLS) and principal component regression (PCR) are the most prominent. Sparse linear regression has been well studied for identifying potential features predictive of drug responses by enforcing elastic net regularization techniques~\citep{Garnett2012,barretina2012cancer,chen2015context}.
	
	Nonlinear drug response analysis including kernel method, neural networks and random forests have also been studied~\citep{sutherland2004comparison,Yamanishi2012,menden2013machine,cortes2015improved, zhang2015predicting, cichonska2015identification}.\\
	In particular, \cite{costello2014community} proposed Bayesian multi-task multiple kernel learning (BMTMKL) to predict drug responses on new human breast cancer cell lines. The BMTMKL method uses a kernelized regression approach that combines multi-task and multi-view learning (i.e., learning from multiple side-data views) with Bayesian inference to estimate the model parameters.  Their results showed that modeling nonlinearities in the data was an essential attribute to predict drug responses. However, the model makes the simplifying assumption that the predictions are based on a single underlying component.
	
	Alternatively, matrix factorization models integrating side-data views have also been studied in drug response analysis. The main idea behind these methods is to jointly factorize the side-data views and output matrix to finding a better low-dimensional latent representation (components) for both rows and columns of the output matrix. To this end,~\cite{zhou2012kernelized} proposed kernelized probabilistic matrix factorization (KPMF), a low-rank matrix factorization method that uses Gaussian process priors with covariance matrix on side-data view. While the method can explain tasks with multiple components, it is, however, limited to a single kernel for each side and, therefore, is unable to learn from multiple side-data views.
	
	Recently, a kernelized Bayesian matrix factorization (KBMF) extending kernelized matrix factorization with fully Bayesian inference, combining multiple side-data views to jointly factorize the output matrix has been proposed~\citep{gonen2013kernelized,gonen2014kernelized}. With side-data views encoded as kernel functions, the main idea is to project each kernel onto a low-dimensional component space, where they are combined with the kernel weights to get a composite component space of the output matrix. The KBMF method has been studied in various applications ranging from drug-target to drug response predictions ~\citep{gonen2012predicting,ammad2014integrative}. However, KBMF integrates multiple side-data views assuming that a source is either relevant for all tasks or none, failing to identify component-specific dependencies between the side-data views and the output matrix.
	\section{Methods}\label{sec:methods}
	\subsection*{Kernelized Bayesian Matrix Factorization ({\it cw}KBMF)}
	We introduce a novel extension of the state-of-the-art kernelized Bayesian matrix factorization method to model the complex associations between a large number of side-data views and the latent component space of the output matrix. This new formulation of kernelized Bayesian matrix factorization (KBMF) allows component-wise multiple kernel learning (MKL), referred to as {\it cw}KBMF for brevity. {\it cw}KBMF is characterized by the ability to comprehensively model the associations that allow two advancements: i) improve the predictive power of the model; and ii) identify the component-specific latent dependencies for interpreting the associations. 
	
	The model is defined for the factorization of a given matrix $\mat{Y} \in \mathbb{R}^{N_{\texttt{x}} \times N_{\texttt{z}}}$, using known sets of $P_x$ side-data views for the rows and $P_z$ side-data views for the columns. In order to represent non-linear associations, similarities between samples in the side-data views are encoded as input kernel matrices $\{\mat{K}_{\texttt{x},m} \in \mathbb{R}^{N_{\texttt{x}} \times N_{\texttt{x}}}\}_{m = 1}^{P_{\texttt{x}}}$ and $\{\mat{K}_{\texttt{z},n} \in \mathbb{R}^{N_{\texttt{z}} \times N_{\texttt{z}}}\}_{n = 1}^{P_{\texttt{z}}}$. Here matrices are denoted by capital letters, with the subscript (${_x}$ or ${_z}$) indicating the corresponding side of the model. All equations are formulated, however, with corresponding scaler entities denoted by non-capital letters, with the superscript denoting the row index and the last subscript representing the column index (i.e., $a_{\texttt{x},s}^{i}$ denotes the entry at (row i, column s) of matrix $\A_{x}$). Without compromising the generalizability, the rest of this article focuses on multiple side-data views in the rows only.
	
	The model is specified as a low-rank factorization of the matrix $\mat{Y}$ such that the latent representations $\mat{H}_x \in \mathbb{R}^{N_{\texttt{x}} \times R}$ and $\mat{H}_z \in \mathbb{R}^{N_{\texttt{z}} \times R}$ are learned jointly from $\mat{Y}$ and the $\mat{K}_{\texttt{x},m}$, $\mat{K}_{\texttt{z},m}$ side-data views. This is achieved by an interplay of two elements. First, each of the  $\{\mat{K}_{\texttt{x},m}\}_{m = 1}^{P_{\texttt{x}}}$ kernels is transformed to a lower dimensional sub-space $\{\mat{G}_{\texttt{x},m} \in \mathbb{R}^{N_{\texttt{x}} \times R}\}_{m = 1}^{P_{\texttt{x}}}$ through a common projection matrix $\mat{A}_x \in \mathbb{R}^{N_{\texttt{x}} \times R}$. The low-rank transformations of the kernels are combined using multiple kernel learning to compute the latent matrix factors $\mat{H}_x$. 
	
	The {\it cw}KBMF method is formulated in a Bayesian setting using conjugate priors, where $\mathcal{N}(\cdot; \mu, \mat{\Sigma})$ denotes the normal distribution with mean $\mu$ and covariance $\mat{\Sigma}$, while $\mathcal{G}(\cdot; \alpha, \beta)$ is the gamma distribution with the parameters, shape $\alpha$ and scale $\beta$. The matrix factorization is formulated as
	\begin{alignat*}{4}
	y_{j}^{i} | h_{\texttt{x},i}, h_{\texttt{z},j} &\sim \mathcal{N}(y_{j}^{i}; h_{\texttt{x},i}^{\top} h_{\texttt{z},j}, \sigma_{y}^{2}) &&\smallspace \forall (i, j)
	\end{alignat*}
	\noindent where $i=1:N_x$ and $j=1:N_z$ denote the samples and $\sigma_{y}$ the noise. Here $h_{\texttt{x},i}$ and $h_{\texttt{z},j}$ are vectors of length $R$, the number of components, and represent the low-dimensional factors of the samples in $Y$.
	
	Our extension formulates this factorization with the novel component-wise MKL and has the distributional assumptions: 
	\begin{alignat*}{4}
	\eta_{\texttt{x},m}^{s} &\sim \mathcal{G}(\eta_{\texttt{x},m}^{s}; \alpha_{\eta}, \beta_{\eta}) &&\smallspace \forall (m, s) \\\
	e_{\texttt{x},m}^{s} | \eta_{\texttt{x},m}^{s} &\sim \mathcal{N}(e_{\texttt{x},m}^{s}; 0, (\eta_{\texttt{x},m}^{s})^{-1}) &&\smallspace \forall (m, s) \\
	h_{\texttt{x},i}^{s} | \{e_{\texttt{x}, m}^{s}, g_{\texttt{x},m,i}^{s}\}_{m = 1}^{P_{\texttt{x}}} &\sim \mathcal{N}\left(h_{\texttt{x},i}^{s}; \sum \limits_{m = 1}^{P_{\texttt{x}}} e_{\texttt{x},m}^{s} g_{\texttt{x},m,i}^{s}, \sigma_{h}^{2}\right) &&\smallspace \forall (s, i) 
	\end{alignat*}
	\noindent where superscript $s=1:R$ denotes the components.
	The novel advancement of this formulation is in learning the latent components $\mat{H}_x$ as a combination of kernel-specific components $\{\mat{G}_{\texttt{x},m} \in \mathbb{R}^{N_{\texttt{x}} \times R}\}_{m = 1}^{P_{\texttt{x}}}$ while segregating between kernels that are component-specific and those which are shared across all components. This is achieved by introducing component-specific kernel weights $\mat{e}_{\texttt{x},m}^{s} \in \mathbb{R}^{P_x \times R}$ that control the activity of each kernel in each component. This extension makes it possible for the method to effectively learn the underlying structure for identifying the associations between kernels and components. The method can also be viewed as combination of component-wise multiple kernel learning and matrix factorization. The $\eta_{\texttt{x},m}^{s}$ defines an element-wise prior for the kernel-weights $e_{\texttt{x},m}^{s}$, making it possible to effectively switch off some of the weights in a component-wise fashion.
	
	Finally, the dimensionality reduction of the model has the distributional assumptions: 
	\begin{alignat*}{4}
	\lambda_{\texttt{x},s}^{i} &\sim \mathcal{G}(\lambda_{\texttt{x},s}^{i}; \alpha_{\lambda}, \beta_{\lambda}) &&\smallspace \forall (i, s) \\
	a_{\texttt{x},s}^{i} | \lambda_{\texttt{x},s}^{i} &\sim \mathcal{N}(a_{\texttt{x},s}^{i}; 0, (\lambda_{\texttt{x},s}^{i})^{-1}) &&\smallspace \forall (i, s) \\
	g_{\texttt{x},m,i}^{s} | {a}_{\texttt{x},s}, {k}_{\texttt{x},m,i} &\sim \mathcal{N}(g_{\texttt{x},m,i}^{s}; a_{\texttt{x},s}^{\top} k_{\texttt{x},m,i}, \sigma_{g}^{2}) &&\smallspace \forall (m, s, i) 
	\end{alignat*}
	\noindent where a joint $A_x$ matrix projects each of the kernels to a low-dimensional representation. The hyper-parameters $\alpha_{\lambda},\beta_{\lambda},\alpha_{\eta},\beta_{\eta},$ $\sigma_{g},\sigma_{h},\sigma_{y}$ can be used to express prior knowledge about the data-generating process, or set to uninformative values (as in this paper).
	
	
	The model is formulated with conjugate priors and variational approximation is used to perform model inference. The computation-al complexity of the model is $\mathcal{O}(R \max(N_{\texttt{x}}^{3}, N_{\texttt{z}}^{3}) + R \max(P_{\texttt{x}}^{3}, P_{\texttt{z}}^{3}))$ which is faster than standard pair-wise kernel approaches \citep{ben2005kernel} and slower only linearly with a factor of $R$ in $\max(P_{\texttt{x}}^{3}, P_{\texttt{z}}^{3})$, in comparison to original KBMF formulation. The model achieves a run time to the tune of minutes for reasonably sized data sets ($\approx5$ minutes of wall clock time on a standard computer, for a single cross validation fold on the largest data studied in this manuscript).
	
	\subsection*{Publicly Available Data sets and Preprocessing}\label{sec:data}
	We used two publicly available cancer data sets to model drug response associations in this study.
	
	\paragraph{{\textbf{Genomics of Drug Sensitivity in Cancer:}}}
	The first data come from Genomics of Drug Sensitivity in Cancer (GDSC) project initiated by Wellcome Trust Sanger Institute version release, June 2014~\citep{yang2013genomics}.
	The data comprised of 124 human cancer cell lines and 124 anti-cancer drugs, for which complete drug response measurements are available and the response range is consistent with earlier publications~\citep{Garnett2012,menden2013machine}. Drug response measurements are summarized as log IC$_{50}$ values (micro molar concentration of a drug required to inhibit 50\% of the cell growth) obtained by curve fitting through the 9-point dose response data. The cell lines are annotated with tissue type,  and drugs with their primary therapeutic targets.
	
	\paragraph{{\textbf{Cancer Therapeutic Response Portal:}}}
	The second data originate from Cancer Theraupetic Response Portal (CTRP) version release v1 2013,~\citep{basu2013interactive} by Broad Institute summarizing area-under-concentration-response curve (AUC) values from 8-point dose response data measured on human cancer cell lines. For our case study, we focus on the set of 66 cell lines and 63 anti-cancer drugs, whose AUC values were observed without missing values. The molecular profiles for the cell lines was obtained from Cancer Cell line Encylopedia CCLE~\citep{barretina2012cancer}. As in GDSC, the cell lines are annotated with tissue type and gene expression, while drugs with their primary therapeutic targets. 
	
	As the input data, we used the baseline gene expression values of all the cell lines quantizing the number of transcripts expressed in a cell. These measurements characterize the genome-wide molecular profiles that may be indicative of the response patterns.
	\paragraph{{\textbf{Prior Biological Knowledge:}}}
	In order to incorporate prior biological knowledge, we used a selected set of pathways and gene sets from Molecular Signature database MSigDB~\citep{liberzon2011molecular}. Specifically, we  extracted the C2CP and C6 collections of pathways and genesets from MSigDB, respectively. C2CP contains pathways compiled from online pathway databases, biomedical literature, published mammalian gene expression studies and MYC target gene database. C6 gene sets denote oncogene signatures of cellular pathways which are often dis-regulated in cancer. These oncogene signatures are computed using microarray data from NCBI GEO and from profiling experiments involving perturbation of known cancer genes. For simplicity in the rest of the paper, we use a common term for both of the collections: {\it pathways}. 
	\subsection*{Experimental Setup}\label{sec:exp_setting}
	\paragraph{{\textbf{Incorporating Prior Biological Knowledge:}}}
	We focused the analyses on drug targets by, for each of the two collections, carefully selecting the subset of pathways that were directly linked to the known primary targets of the drugs. This was done by examining the correspondence between pathway names and the known primary targets of the drugs. The drug target data coming from the original annotations of GDSC and CTRP was used for this purpose. The gene expression data were then split into groups of genes, where each group represented one pathway. All the other genes which were not part of any of the target-based pathway selection, were collected in a separate single group (collectively called as {\it ``other genes''}). When the variable groups in the data are constructed in this way, the component-wise MKL based data integration can choose what prior knowledge is useful for predicting responses. Still, no knowledge is lost as all variables have been included, and additionally allows to learn associations between other genes and the responses. The total number of groups formed per case study are listed in Table~\ref{tab:datasets}. We term each group with a keyword {\it `view'}.
	
	\begin{table}[h]
		\caption{Data used in the drug response predictions.}\label{tab:datasets}
		\resizebox{12.5cm}{!}{
			\begin{tabular}{llllll}
				\hline
				Datasets & Cells & Drugs & Genes & Primary Targets & Views\\
				\hline
				GDSC & 124 & 124 & 13321 & 60 & 72 (71 pathways, 1 other genes) \\
				CTRP & 66 & 63 & 18988 & 58 & 26 (25 pathways, 1 other genes) \\
				\hline
			\end{tabular}}
		\end{table}
		Additional information about the data including the names of the cell lines, drugs, primary targets and pathways can be found from the supplementary material. The response data consist of both types of drugs: FDA approved `drugs' and `investigational chemical compounds'. In the paper, we use both of these terms interchangeably.
		\paragraph{{\textbf{Cross-Validation:}}}
		We compared the performance of {\it cw}KBMF with several methods including KBMF$_\textnormal{multi-view}$; kernelized Bayesian matrix factorization with pathway based groups, BMTMKL$_\textnormal{multi-view}$; Bayesian multi-task learning with pathway based groups, KPMF$_\textnormal{single-view}$; kernelized probabilistic matrix factorization without pathway based groups, MT-LR$_\textnormal{single-view}$; multi-task sparse linear regression without pathway based groups and the classical Baseline; mean of the training drug response data (assuming no genomic data is available). 
		
		We performed a 5-fold cross validation procedure, where in each fold a randomly selected subset of cell lines is completely held-out (as test cell lines) and models were trained on the remaining cell lines (training data). To establish robust findings the 5-fold cross validation procedure was repeated 10 times with different random cross-validation folds.
		
		For the kernelized Bayeisan methods (BMTMKL, KBMF and {\it cw}KBMF), we use uninformative priors for the projection matrices and the kernel weights. In particular,  the hyperparameter values for BMTMKL are selected as ($\alpha_{\lambda}$, $\beta_{\lambda}$,$\alpha_{\upsilon}$, $\beta_{\upsilon}$, $\alpha_{\gamma}$, $\beta_{\gamma}$, $\alpha_{\omega}$, $\beta_{\omega}$, $\alpha_{\epsilon}$, $\beta_{\epsilon}$) = (1, 1, 1, 1, 1, 1, 1, 1, 1, 1) and for KBMF, {\it cw}KBMF are selected as ($\alpha_{\eta}$, $\beta_{\eta}$, $\alpha_{\lambda}$, $\beta_{\lambda}$) = (1, 1, 1, 1), and the standard deviations ($\sigma_{g}$, $\sigma_{h}$, $\sigma_{y}$) are set to (0.1, 0.1,1). For KPMF, the standard deviation $\sigma_{y}$ is set to one. For the side-data views, we computed Gaussian kernels, where the width parameter $\sigma$ was set in the standard way ($\sigma = $ dimensionality of the side-data view). The drug response measurements were normalized to have zero mean and unit variance.
		
		We used multi-task sparse linear regression using the {\it glmnet} package~\citep{friedman2010regularization}. The sparse linear regression has two parameters that are to be optimized: $\alpha$ (elastic net mixing parameter) and $\lambda$ (the penalty parameter). For each test set prediction, we performed a nested 5-fold cross validation procedure on the training data, to choose optimal values for $\alpha \in [0,1]$ with an increment of 0.1 and $\lambda$ (from 100 values). We finally selected a combination of $\alpha$ and $\lambda$ values that gave minimum error averaged over the cross-validated folds.
		\paragraph{{\textbf{Evaluation Criteria:}}}
		We evaluated the predictive performance of {\it cw}KBMF and other methods using drug-wise spearman correlation as an evaluation criterion and report an averaged correlation for each drug from 10 random repeats of the cross-validation procedure. In addition, the correlations were averaged to obtain a cumulative correlation value for each method.
		\section{Results and Discussion}\label{sec:res-discussion}
		\subsection*{Synthetic Data Set}
		\begin{figure*}[!ht]
			\centering
			\includegraphics[width=1\textwidth,height=1\textheight,,keepaspectratio]{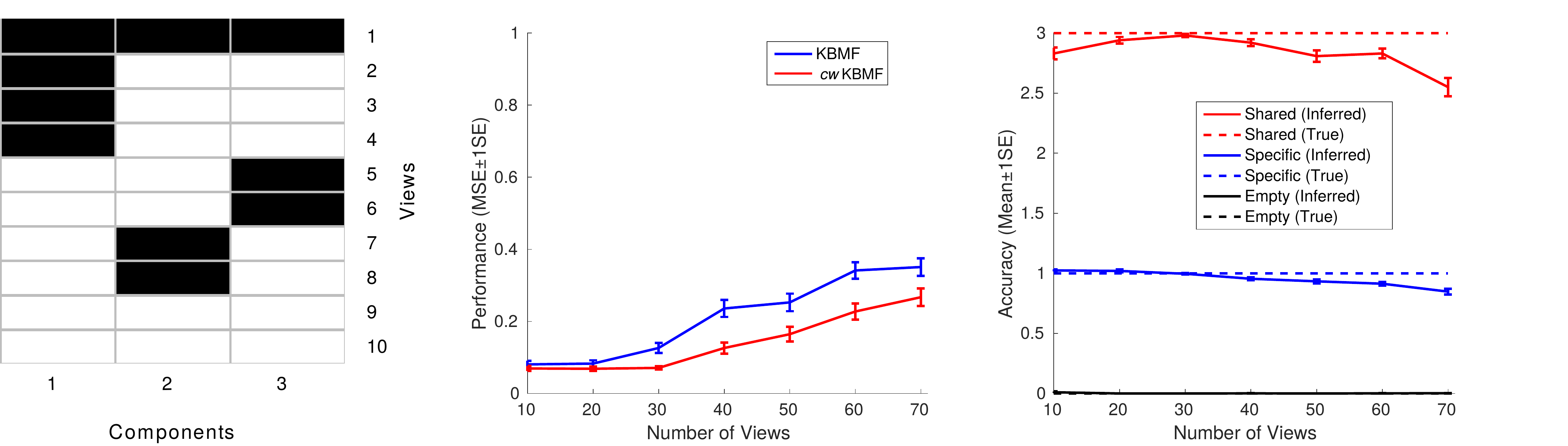}
			\caption{Identification of component-view activities and predictions on synthetic data set. Method abbreviation: {\it cw}KBMF, kernelized Bayesian matrix factorization with component-wise MKL; KBMF, kernelized Bayesian matrix factorization. Left: The component-view activities learned by the model. Black indicates that a view is active in a component while white represents not-active. Middle: the mean squared error (MSE) of predictions, averaged over 100 data sets at each point (and bars denoting 1-standard error of the mean (1SE)). The performance is indicative of the models ability to discover the underlying structure of the data. Right: The accuracy of the model to discover component-view associations. The true and the inferred, averaged accuracy of the associations from 100 data sets are marked for shared, specific and empty component, along with 1SE. The figure demonstrates the models ability to accurately discover the component-view associations.}
			\label{fig:syn_pred}
		\end{figure*}
		To demonstrate that the model can infer the true associations between multiple side-data views and components, we perform the first experiment using synthetic data set. Specifically, the {\it cw}KBMF method has been designed to learn the complex relationships patterns, by representing them as activity profiles of components over the views.
		
		To this end, 100 synthetic data sets $\Y$ with ${N_{\texttt{x}}=100, N_{\texttt{z}}=100}$ and $R=3$ components were generated such that each data set was supplemented with $P_x=10$ side-data views (encoded as kernels). 
		The associations between the $P_x=10$ side-data views and $K=3$ components were encoded such that one view is active in all the components (shared), while the rest are equally split into K+1 sets, where each view is either active in one component (specific) or not active in any of the components (empty). Here the key assumption is that given the kernels for $P_x=10$ side-data views and the output matrix $\Y$, the model decomposes $\Y$ into components while accurately learning the associations between kernels and components. In addition, $1\%$ values in each $\Y$ were marked as missing data (test set) to measure the predictive accuracy of the model.
		
		The model is run for each of the 100 data sets to learn the associations. The component-view weights $e_x^s$ represent the activity of each view $x=1:P_x$ in components $s=1:R$. Since the model is encoded with an element-wise prior it can be effectively thresholded to illustrate component-view activity. In order to focus on the most important associations for each component, we consider the associations that are notably strong with respect to the prior (i.e., z-score ($e_x^s$) $>$ 0.67) as active. Figure \ref{fig:syn_pred} (left) shows the resulting component-view activities inferred by the model for $P_x=10$. The figure demonstrates that the model is able to accurately discover the component-view activities as inserted in the data (described above), up to a random permutation of the components.
		
		Next, we measured the accuracy of the model in inferring the component-view activities as well as in predicting unobserved values in $Y$ over a range of side-data views $P_x$. The associations were learned and prediction performance was evaluated for 100 data sets for each value of $P_x$. Figure~\ref{fig:syn_pred} (right panel) demonstrates the accuracy of learning the associations, particularly the model performs well in discovering the shared, specific as well as empty components over the range of input views. In addition, Figure 1 (middle panel), {\it cw}KBMF consistently outperforms KBMF in the prediction task as well, especially when the number of views is large. As expected, the performance of the methods deteriorates as the number of views (dimensionality) increases. However, {\it cw}KBMF performs reasonably well over the number of views applicable to the drug-response prediction data sets in this study.
		
		\subsection*{Cancer Data Sets}
		We next compare {\it cw}KBMF with alternatives on two  case studies GDSC and CTRP, and report their predictive performance in the 5-fold cross validation procedure (described in Section~\ref{sec:exp_setting}). 
		To evaluate the new model extension and the benefit of the principled incorporation of prior knowledge, we compare {\it cw}KBMF's performance to other methods in two scenarios, 
		\vspace{-2.5mm}
		\begin{enumerate}
			\item Genomic Data + Prior Knowledge: The genomic features are divided into several side-data views based on the prior knowledge about the pathways. We represent this scenario with a subscript $multi-view$ in the results; and
			\item Genomic Data (only): The genomic features are used as a single view and does not benefits from the prior knowledge. We denote this scenario with a subscript $single-view$ in the results.
		\end{enumerate} 
		\vspace{-2.5mm}
		\begin{figure*}
			\centering
			\includegraphics[width=1\textwidth,height=1\textheight,keepaspectratio]{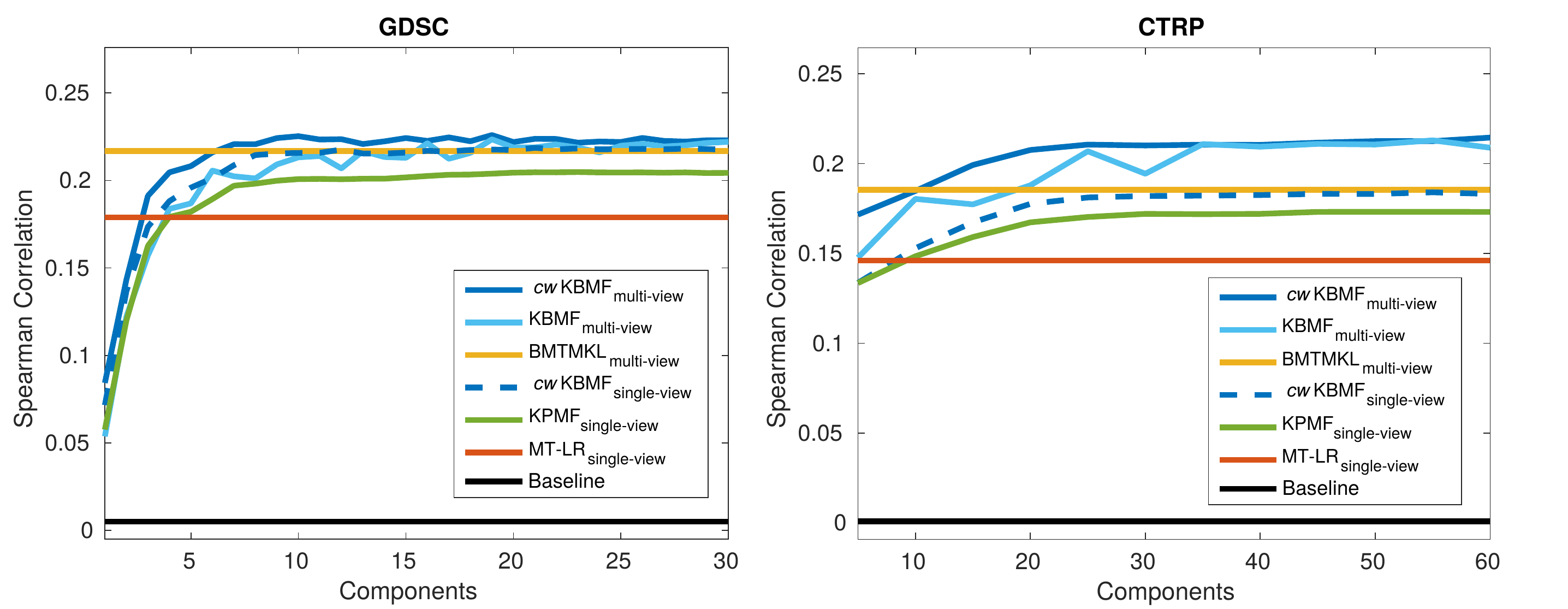}
			\caption{Prediction performances (Spearman correlation) averaged over drugs with a 5 fold cross-validation procedure repeated 10 times. GDSC data set (left) and CTRP data set (right). Method abbreviation: {\it cw}KBMF, kernelized Bayesian matrix factorization with component-wise MKL; KBMF, kernelized Bayesian matrix factorization; BMTMKL, Bayesian multi-task MKL; KPMF, kernelized probabilistic matrix factorization; MT-LR, multi-task sparse linear regression; Baseline, mean of the training data. The predictive performance obtained by {\it cw}KBMF for both scenarios is found to be significantly higher than the others (p$<$0.05; one-sided paired Wilcoxon Sign-Rank test corrected for multiple testing, supplementary material Tables S3 and S6).}
			\label{fig:sanger_ctrp_cv_pred}
		\end{figure*}
		Figures \ref{fig:sanger_ctrp_cv_pred} show the predictive performances of all the methods on the GDSC (left) and CTRP (right) data sets. {\it cw}KBMF outperforms its competitors for both scenarios. Even though the differences in performances are rather small, the predictive performance obtained by {\it cw}KBMF for both scenarios is found to be significantly higher than the others (p$<$0.05; one-sided paired Wilcoxon Sign-Rank test corrected for multiple testing, supplementary material Table S3) in GDSC data set respectively. Similarly in CTRP data set, the predictive performance obtained by {\it cw}KBMF for both scenarios is also significantly higher than the others (p$<$0.05; one-sided paired Wilcoxon Sign-Rank test, corrected for multiple testing, supplementary material Table S6). In the GDSC and CTRP data sets, the maximum predictive performance of {\it cw}KBMF is achieved with 10 and 20 components, respectively. However, in-case of multiple maxima a practical choice could be to prefer solutions with smallest-$R$ in the interest of simpler representations. We chose these components and discuss a detailed comparison of the predictions of {\it cw}KBMF with other methods. 
		
		In GDSC data set, {\it cw}KBMF$_\textnormal{single-view}$ outperforms Baseline, MT-LR$_\textnormal{single-view}$ and KPMF$_\textnormal{single-view}$ (p$<$0.05; one sided paired Wilcoxon Sign-Rank test, for comparing Spearman correlations). {\it cw}KBMF$_\textnormal{multi-view}$ outperforms KBMF$_\textnormal{multi-view}$ and Baseline methods (p$<$0.05; one-sided paired Wilcoxon Sign-Rank test, supplementary material Figure S1). Although {\it cw}KBMF$_\textnormal{multi-view}$ performance is better than BMTMKL$_\textnormal{multi-view}$ (averaged Spearman correlations 0.2253 and 0.2167, respectively), the difference is not statistically significant (p$=$0.11; one-sided paired Wilcoxon Sign-Rank test).
		
		Similarly, in CTRP data set, {\it cw}KBMF$_\textnormal{single-view}$ outperforms Baseline and MT-LR$_\textnormal{single-view}$ (p$<$0.05; one-sided paired Wilcoxon Sign-Rank test). Although {\it cw}KBMF$_\textnormal{single-view}$ performance is better than KPMF$_\textnormal{single-view}$ (averaged Spearman correlation  0.1776 and 0.1673, respectively), the difference is not statistically significant (p$=$0.07, one-sided paired Wilcoxon Sign-Rank test).
		{\it cw}KBMF$_\textnormal{multi-view}$ give better predictions than Baseline, BMTMKL$_\textnormal{multi-view}$ and KBMF$_\textnormal{multi-view}$  (p$<$0.05; one-sided paired Wilcoxon Sign-Rank test, supplementary material Figure S3).\\
		
		The prediction results generalize previous findings that non-linear models improve drug response predictions~\citep{costello2014community}.
		Figure~\ref{fig:sanger_ctrp_cv_pred} clearly shows that non-linear methods are better than the linear counterpart, for predicting drug responses in both data sets.
		
\begin{figure*}
			\begin{center}	\includegraphics[width=1\textwidth,height=1\textheight,keepaspectratio]{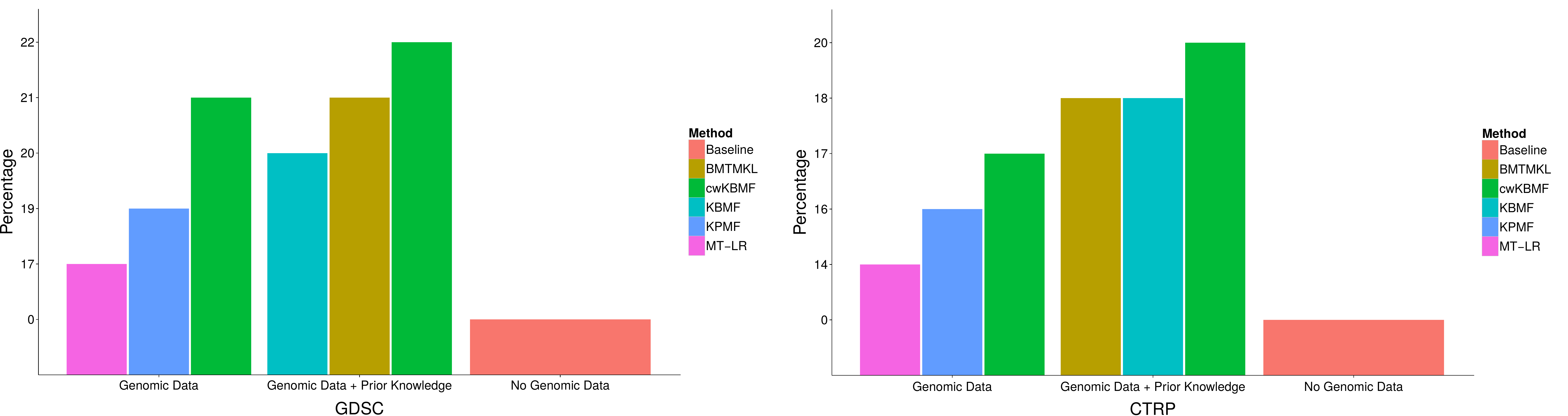}
				\caption{Pathway-based groups of genes (prior biological knowledge) improves predictive performance. Left, GDSC data set and Right, CTRP data set. The height of the bar (y-axis) denotes the percentage increase in performance relative to Baseline, computed using the Spearman correlations averaged over drugs. On x-axis, the bars are grouped based on the type of learning data used, where `` Genomic Data'' means that all of the genes are used as one group and ``Genomic Data $+$ Prior Knowledge'' means that all of the genes are used, grouped into several sets based on the pathway knowledge and lastly ``No Genomic Data'' implies that only mean of the training drug response data is used for prediction.}
				\label{fig:prior-bio-know}
			\end{center}
		\end{figure*}

		Having established that our model outperforms existing methods in both single-view and multi-view settings, we next specifically study the advantage of using prior pathway and target knowledge. To this end, Figure~\ref{fig:prior-bio-know} illustrates the improvement in performance (in \% units) relative to Baseline and when genomic data is supplemented with prior knowledge. 
		
\begin{figure}
			\centering
			\includegraphics[width=0.8\textwidth,height=0.8\textheight,keepaspectratio]{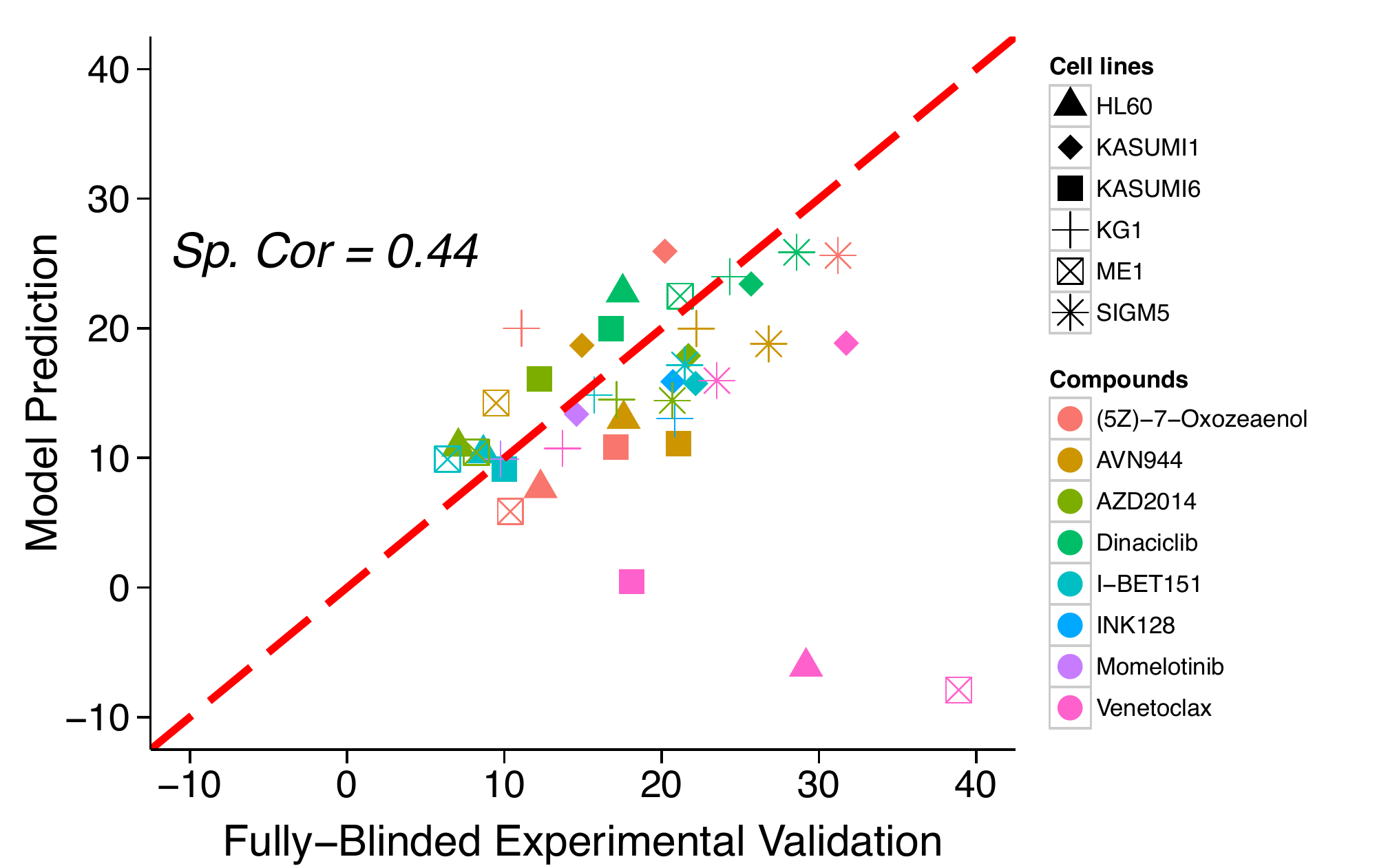}
			\caption{Prediction of the drug sensitivity score (DSS;~\citep{yadav2014quantitative}) of 8 compounds in 6 AML cell lines. The y-axis shows the predictions made by the {\it cw}KBMF model and the x-axis the corresponding validations as measured in the lab. The predictions have a spearman correlation of 0.44, and the correlation increases to 0.70, if the outlier venetoclax is excluded.}
			\label{fig:aml_validation}
		\end{figure}

		As the first observation, the introduction of genomic data via different methods outperforms the baseline predictions demonstrating the genomic features are response predictive. Secondly, incorporating prior biological knowledge improves the prediction performance systematically over a range of methods. Third, systematically modelling the associations between pathway-based genomic profiles and drug response with {\it cw}KBMF outperforms the existing approaches in predicting drug responses. Specifically, in case of the GDSC data set, using genomic data with {\it cw}KBMF improves the prediction performance by 21\% and when the genomic data is supplemented with prior knowledge the performance is improved by 22\%. Similarly, in case of the CTRP dataset, using genomic data with {\it cw}KBMF improves the prediction performance by 17\% and when the genomic data is supplemented with prior knowledge the performance is improved by 20\%. The findings also suggest that incorporating the prior knowledge is more beneficial when the number of samples is smaller (for instance, in the CTRP data set).\\
		\paragraph{\bf Fully Blinded Experimental Validations:}
		Finally, we experiment-ally validated the drug response predictions of our model using an in-house Acute Myeloid Leukemia (AML) cell line panel (Malani et al., manuscript in preparation). The model is learned, analogously to the experiments with public data sets above, using the available training drug response data. Specifically, we made drug response predictions for 8 compounds using 6 AML cell lines of which 83\% measurments were not available for initial model training. To validate the predictions, an independent experiment was carried out in laboratory. The predicted drug responses were found to be correlated with the independent lab measurements (Spearman correlation 0.44~Figure~\ref{fig:aml_validation}, p$<$0.05; compared to the distribution of correlation values obtained via randomization; supplementary material Figure S4). This fully-blinded experimental validation confirms the predictive power of the model, and gives confidence that \emph{in silico} predictions are fairly robust and may be used to study the spectrum of therapeutic choices.
		\subsection*{Inferring Pathway-Drug Response Associations}
		The use of prior knowledge not only improves the prediction performance, but also helps to infer pathway-drug response associations by {\it cw}KBMF, being the first kernelized method making it possible to study such associations. We next study the pathway-drug response associations in the GDSC dataset.
		
\begin{figure*}
			\begin{center}
				\includegraphics[trim=20 80 20 40, scale=0.25,keepaspectratio]{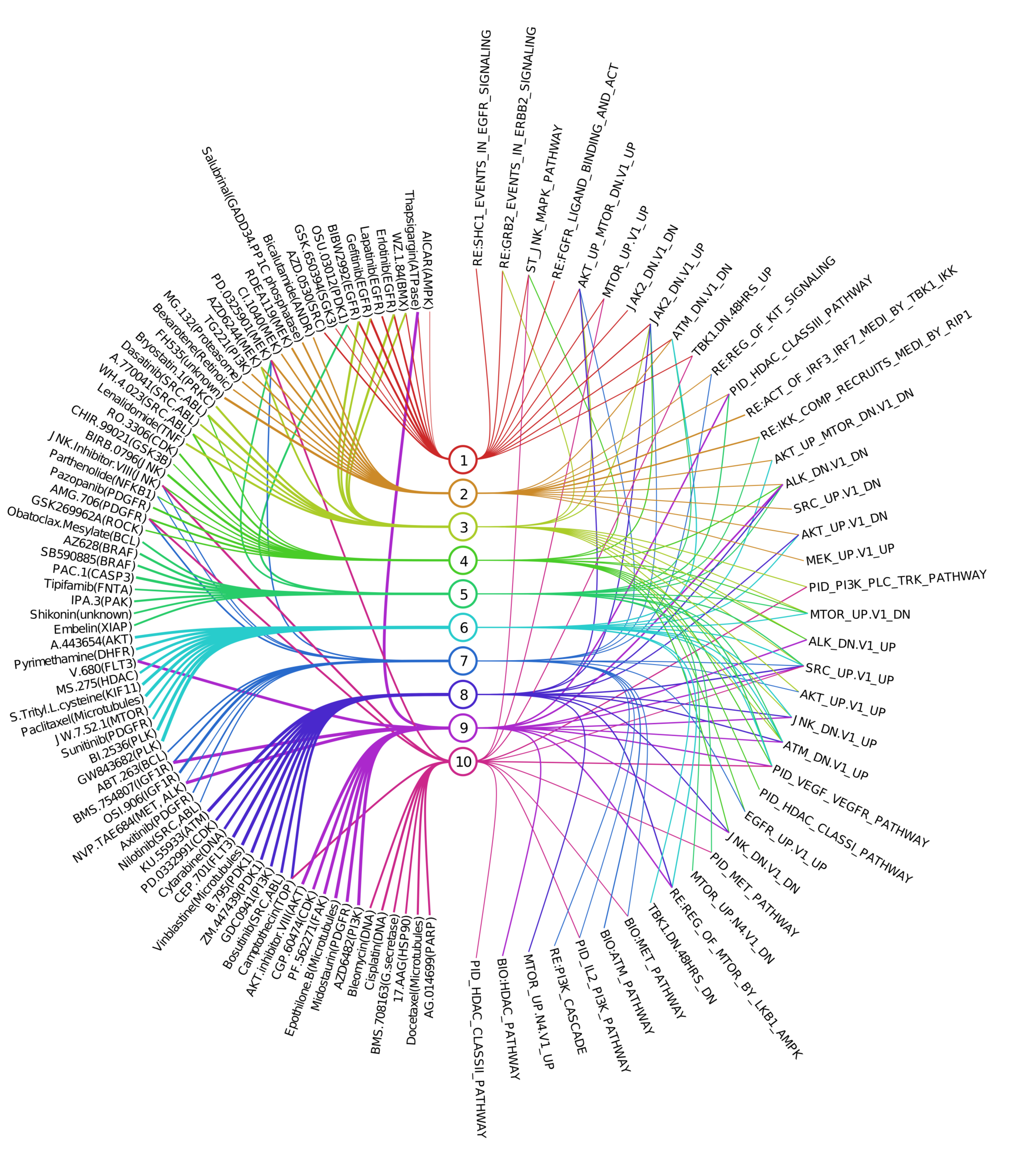}
				\caption{Pathway-response associations decomposed into components in the GDSC data set, visualized as an ``eye diagram'' showing {\it cw}KBMF 10 components (circles), connecting pathways (right) and drugs (with their primary targets in parenthesis; left). The widths of the curves from the components to pathways and drugs indicate the strength of the corresponding associations. For each component, 10 drugs and 10 pathways showing the largest strength are shown.
				}
				\label{fig:sanger_eye}
			\end{center}
		\end{figure*}

		We selected the model learned with 10 components based on cross-validation (as discussed in section~\ref{sec:res-discussion}) and show the pathway-drug response associations in Figure~\ref{fig:sanger_eye}. 
		A component can be characterized by the set of pathways that are active in it and the drugs whose responses they are predictive off, yielding hypotheses of pathways associated with drug responses. The hypotheses generated by all the ten components are illustrated in Figure~\ref{fig:sanger_eye}, while those from the first two components are elaborated in detail below. In order to analyse target-driven effects, the components were sorted based on the consistency of the drug targets in the components.
		\paragraph{Component 1} is characterized by EGFR/ERBB2 inhibitors, lapatinib, erlotinib, BIBW2992 (afatinib) and gefitinib. 
		On the pathway side, we found {\it reactome SHC1 events in EGFR signalling, reactome GRB2 events in ERBB2 signaling}, among the top 10 pathways. It is biologically meaningful that the inhibitors are related to the EGFR signaling, making it possible to inhibit the pathways activity in cancer using the EGFR inhibitors. It is also evident that signaling pathways RAS-RAF-MAPK, PI3K/AKT and JAK/STAT mediate the downstream effect of EGFR autophosphorylation,  thus affecting cellular proliferation, anti-apoptosis, metastasis and tumor invasion~\citep{whirl2012pharmacogenomics}. We give additional details of component 1 explaining the variation of EFGR responses in supplementary material Figure S2 (left). Other drugs explained by the component are aicar (target: AMPK agonist), thapsigargin (target: ATPase, Ca++ transporting, cardiac muscle, slow twitch 2), OSU-03012 (target: PDK1/PDPK1), GSK-650394 (target: SGK3), WZ-1-84 (target: BMX) and AZD-0530 (target: SRC, ABL1). The pathways involved in mediating the downstream signaling may generate novel hypotheses for the action mechanism of these drugs.  
		\paragraph{Component 2} is representative of MEK inhibitors RDEA119 (refametinib), PD-0325901, CI-1040 and AZD6244. Interestingly, on the pathway side, {\it MEK up..v1 up} is identified as one of the top pathways (shown in Figure~\ref{fig:sanger_eye}). It is biologically plausible that the drugs are connected to the up-regulation of MEK pathway, making it possible to inhibit the pathway activity in cancer using the MEK inhibitors. It is also known that MEK inhibition leads to PI3K/AKT activation~\citep{turke2012mek}, supporting the identification of the AKT-related pathways in this component. In general, stimulation of the PI3K/AKT/mTOR cascade enhances growth, survival, and metabolism of many cancer cells, and therefore PI3K/AKT/mTOR signaling pathway is a promising therapeutic target for cancer therapy. We give additional details of component 2 explaining the variation of MEK responses in supplementary material Figure S2 (right). Other drugs explained by the component are bexarotene (target: Retinioic acid X family agonist), bicalutamide (target: Androgen receptor ANDR), MG-132 (target: Proteasome), TGX221 (target: PI3K beta), Salubrinal (target: GADD34-PP1C phosphatase) and FH535. In particular FH535 primary target is unknown, however it has been shown to downregulate the activity of Wnt/$\beta$-Catenin signaling pathway~\citep{gedaly2014targeting,liu2014fh535}. The presence of FH535 in this component suggests potential associations between FH535 response, MEK and AKT-related pathways, which could be further investigated in the lab to identify novel biomarkers for predicting FH535 responses. 
		The analysis conclude that pathway-drug response associations provide biologically meaningful findings. Even though these are well-known cancer-related pathways (serving as proof-of-concept positive controls), the current clinical challenge is to find the patients in which these pathways are perturbed, making it possible to select targeted treatments like MEK inhibitors individually. 
		\section{Conclusion}
		We extended the KBMF method with a novel approach of component-wise MKL. In experiments with two publicly available cancer data sets, the new method showed improved predictive performance compared to other methods (including its predecessor KBMF). Additionally, we confirmed the predictive performance of the method using an in-house AML cell line panel with experimental validation, performed independently in the lab. We also showed that incorporating prior knowledge in the form of pathways helps to improve the prediction performance. We also demonstrated the usefulness of component-wise MKL, combined with prior knowledge for inferring the associations between pathways and drug responses. This way of analyzing drug responses with groups of genes (encoded in the form of pathways) may enhance our understanding of the action mechanism of drugs and can potentially be used to identify novel predictive biomarkers for designing new therapies in cancer. In the future, the method could further be extended with strict sparsity assumptions for component-wise MKL, facilitating the discovery of potentially strong associations between pathways and drug responses. 
		\section*{Acknowledgement}
		\paragraph{Funding:} This work was financially supported by the Academy of Finland (Finnish Center of Excellence in Computational Inference Research COIN; grants  295503 and 292337 to MA and SK;  grants 272437, 269862, 279163 to TA), and Cancer Society of Finland (TA).
		We acknowledge the computational resources provided by Aalto Science-IT project and CSC-IT Center for Science Ltd.
		
		\section*{Availability:}
		The source code implementing the method is available at \url{http://research.cs.aalto.fi/pml/software/cwkbmf/}.
		\bibliographystyle{apa}
		\bibliography{cwkbmf.bib}

\newcommand{\beginsupplement}{%
	\setcounter{section}{0}
	\renewcommand{\thesection}{S\arabic{section}}%
	\setcounter{subsection}{0}
	\renewcommand{\thesubsection}{S\arabic{subsection}}%
	\setcounter{table}{0}
	\renewcommand{\thetable}{S\arabic{table}}%
	\setcounter{figure}{0}
	\renewcommand{\thefigure}{S\arabic{figure}}%
}
\beginsupplement
\section{Supplementary Material}\label{sec:supp-mat}

\subsection*{Case Study: GDSC Data Set}
\begin{figure*}[]
\begin{center}
	\includegraphics[width=1\textwidth,height=0.5\textheight,keepaspectratio]{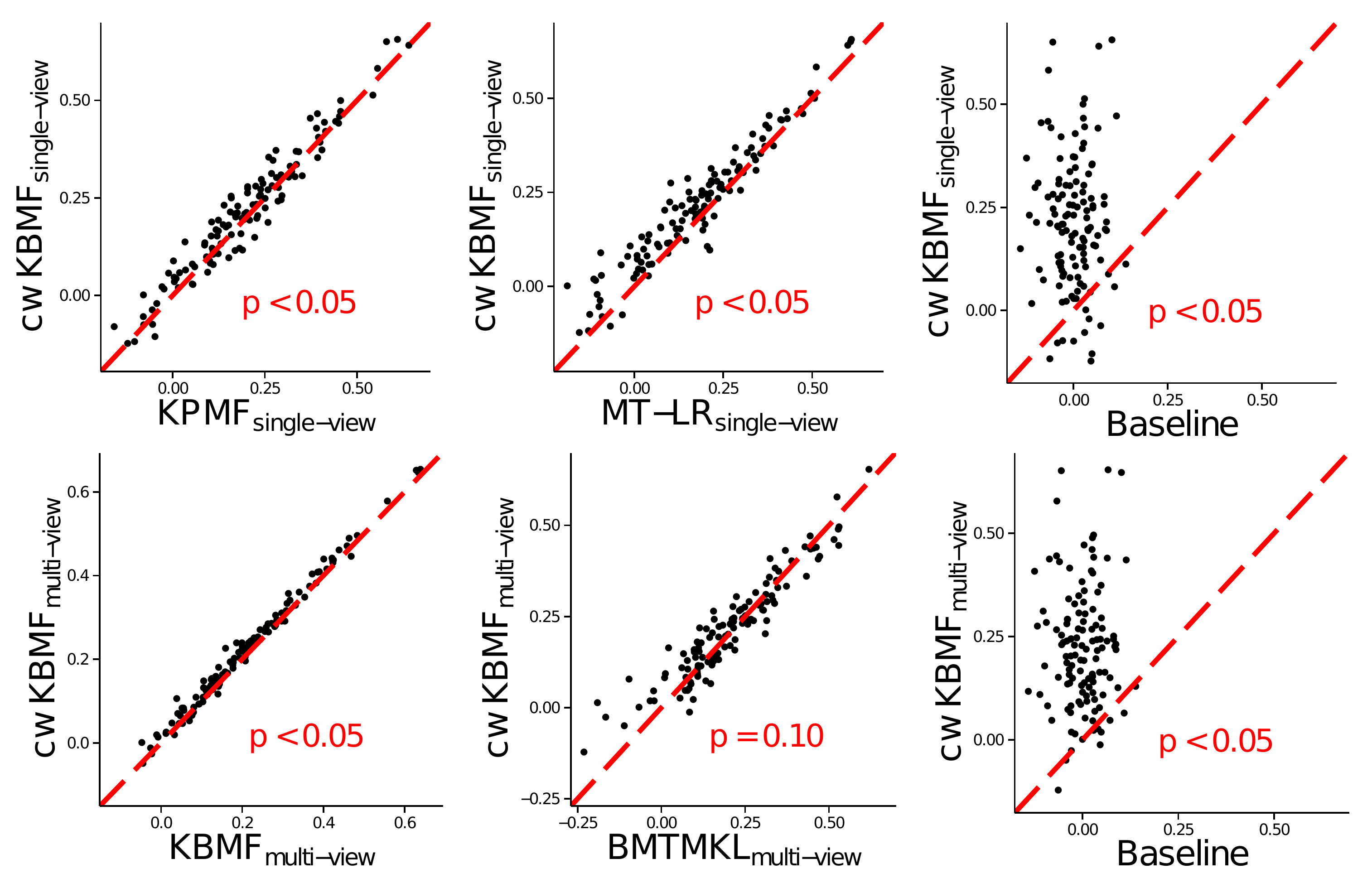}
	\caption{Comparison of {\it cw}KBMF with other methods on GDSC data. Spearman rank correlation between the predictions and the response data, for each drug are calculated for each method. The correlations from each method (x-axis) are compared to those from {\it cw}KBMF (y-axis) with components 10. Each dot represents cross validated prediction performance for a drug response averaged over 10 random rounds of 5 fold cross-validation procedure. The subscript in method's name denotes the format of learning data used by the methods; {\it single-view} means that the method is learned using one view (i.e, one group containing all the genomic features), while {\it multi-view} means that the method is learned using multiple views (i.e., genomic features are grouped into many views based on the prior knowledge about the pathways).
	Method abbreviation: {\it cw}KBMF, kernelized Bayesian matrix factorization with component-wise MKL; KBMF, kernelized Bayesian matrix factorization; BMTMKL, Bayesian multi-task MKL; KPMF, kernelized probabilistic matrix factorization; MT-LR, multi-task sparse linear regression; Baseline, mean of the training data. P-values show the statistical significance of the improvements of {\it cw}KBMF’s predictions compared to other methods computed with one-sided paired Wilcoxon Sign-Rank test.}
	\label{fig:sanger_pred_k}. 
\end{center}
\end{figure*}

\begin{figure}
\centering
\begin{subfigure}{.5\textwidth}
  \centering
  \includegraphics[width=0.9\linewidth]{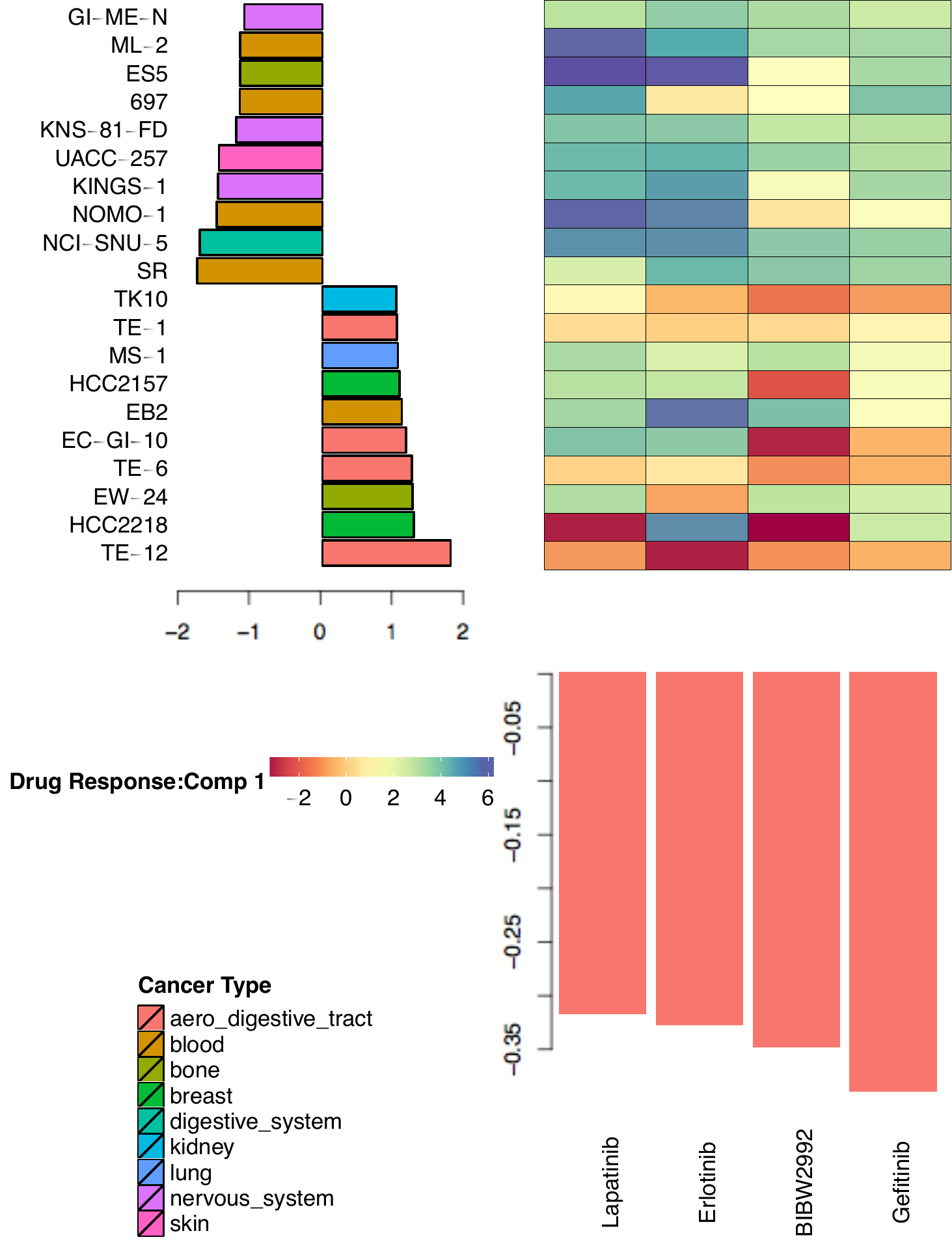}
  \caption{Component 1}
  \label{fig:comp1}
\end{subfigure}%
\begin{subfigure}{.5\textwidth}
  \centering
  \includegraphics[width=0.9\linewidth]{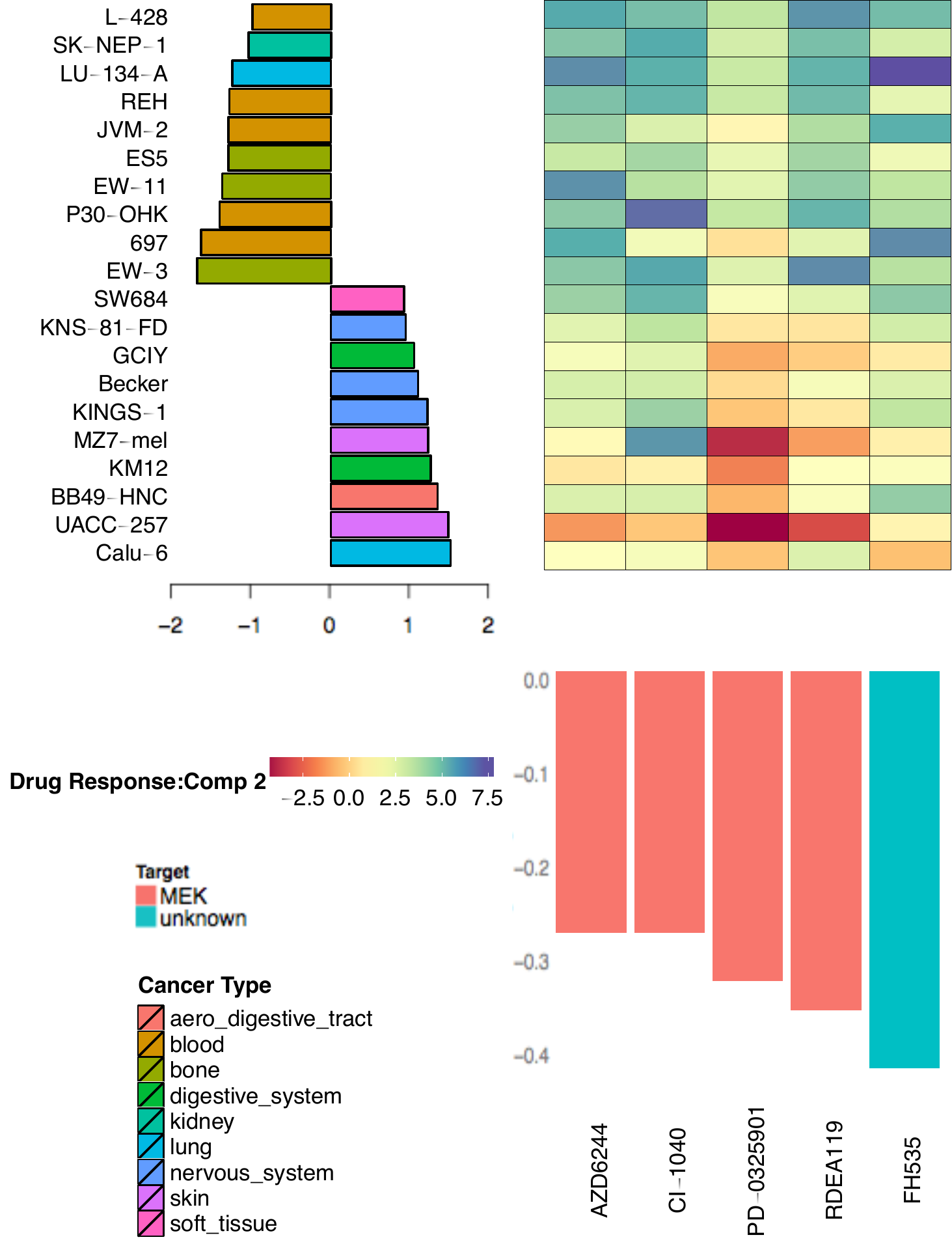}
  \caption{Component 2}
  \label{fig:comp2}
\end{subfigure}
\caption{Variation in drug responses explained by {\it cw}KBMF components. Component 1 explains EGFR inhibitors and component 2 models MEK inhibitors. The heat map (in center) shows the actual drug responses of the inhibitors (x-axis), on top 10 cancer cell lines (y.axis). Bar plots on x- and y- axis, indicate the weights of the inhibitors and cell lines learned by the {\it cw}KBMF method, respectively. FH535 (with unknown primary target) explained by component 2, shows variations in responses similar to MEK inhibitors.}
\label{fig:sanger_comp1_comp2}
\end{figure}

Below, we provide supplementary information of the data and results from the GDSC data set as discussed in the main text sections 3 and 4.\\
Table~\ref{SI-tab-GDSC-Dataset} shows the names, type and tissue of 124 cancer cell lines and names, primary targets of 124 drugs used in our study. We obtained these data from Genomics of Drug Sensitivity in Cancer project~\cite{yang2013genomics}, and show here for the purpose of motivating reproducibility of the results. For defining compact primary targets, we annotated their names so that they belong to one class of targets, for instance MEK1/2 were named to MEK. Table~\ref{SI-tab-GDSC-Pathways} list the 71 pathways used in the case study.\\
Table~\ref{tab:pvalues-sanger_cv_pred} shows the statistical significance of the predictive performances of {\it cw}KBMF compared to other methods on the GDSC data set. Figure~\ref{fig:sanger_pred_k} compares the predictive performance of {\it cw}KBMF with other methods on GDSC data, when learned with 10 components (see details in main text section 4). \\
Figure~\ref{fig:sanger_comp1_comp2} shows the variation of drug responses explained by the two components of the {\it cw}KBMF model. 
For component 1, the drug response pattern can be hypothesized in the following manner: the EGFR inhibitors showing a response to the set of cell lines (at the lower part), inhibit the EGFR activity which in turn inhibits the activation of the downstream signaling pathways, whereas the inhibitors showing no response to the cell lines (on the upper part) fail to inhibit the EGFR activation, resulting the downstream signaling pathways to be active. \\
Similarly for component 2, the MEK inhibitors showing response to the set of cell lines (at the lower part), inhibit the MEK activity,, whereas the inhibitors showing no response to the cell lines (on the upper part) fail to inhibit the MEK pathways.
\begin{table}[]
\centering
\resizebox{\textwidth}{!}{%
\begin{tabular}{@{}lllllll@{}}
\hline
 & Cell line & Type & Tissue & Drug & Primary Target & Annotated Target \\ 
\hline
1 & ES3 & bone & ewings\_sarcoma & Erlotinib & EGFR & EGFR \\
2 & ES5 & bone & ewings\_sarcoma & Rapamycin & MTOR & MTOR \\
3 & EW-11 & bone & ewings\_sarcoma & Sunitinib & PDGFRA, PDGFRB, KDR, KIT, FLT3 & PDGFR \\
4 & NCI-H1395 & lung & lung\_NSCLC\_adenocarcinoma & PHA-665752 & MET & MET \\
5 & NCI-H1770 & lung & lung\_NSCLC\_not\_specified & MG-132 & Proteasome & Proteasome \\
6 & DMS-114 & lung & lung\_small\_cell\_carcinoma & Paclitaxel & Microtubules & Microtubules \\
7 & NCI-H1092 & lung & lung\_small\_cell\_carcinoma & Cyclopamine & SMO & SMO \\
8 & NCI-H2141 & lung & lung\_small\_cell\_carcinoma & AZ628 & BRAF & BRAF \\
9 & NCI-H345 & lung & lung\_small\_cell\_carcinoma & Sorafenib & PDGFRA, PDGFRB, KDR, KIT, FLT3 & PDGFR \\
10 & NCI-H446 & lung & lung\_small\_cell\_carcinoma & VX-680 & Aurora A/B/C, FLT3, ABL1, JAK2, & FLT3 \\
11 & NCI-H82 & lung & lung\_small\_cell\_carcinoma & Imatinib & ABL, KIT, PDGFR & PDGFR \\
12 & SK-N-DZ & nervous\_system & neuroblastoma & NVP-TAE684 & ALK & MET, ALK \\
13 & Calu-6 & lung & lung\_NSCLC\_adenocarcinoma & PF-02341066 & MET, ALK & MET, ALK \\
14 & LU-65 & lung & lung\_NSCLC\_large\_cell & AZD-0530 & SRC, ABL1 & SRC \\
15 & SHP-77 & lung & lung\_small\_cell\_carcinoma & S-Trityl-L-cysteine & KIF11 & KIF11 \\
16 & HCC1187 & breast & breast & Z-LLNle-CHO & g-secretase & g-secretase \\
17 & HCC2157 & breast & breast & Dasatinib & ABL, SRC, KIT, PDGFR & SRC,ABL \\
18 & HCC2218 & breast & breast & GNF-2 & BCR-ABL & SRC,ABL \\
19 & BB49-HNC & aero\_digestive\_tract & head\_and\_neck & CGP-60474 & CDK1/2/5/7/9 & CDK \\
20 & CPC-N & lung & lung\_small\_cell\_carcinoma & CGP-082996 & CDK4 & CDK \\
21 & EC-GI-10 & aero\_digestive\_tract & oesophagus & A-770041 & SRC family & SRC,ABL \\
22 & IM-9 & blood & Myeloma & WH-4-023 & SRC family, ABL & SRC,ABL \\
23 & IST-SL1 & lung & lung\_small\_cell\_carcinoma & WZ-1-84 & BMX & BMX \\
24 & LB831-BLC & urogenital\_system & Bladder & BI-2536 & PLK1/2/3 & PLK \\
25 & LU-134-A & lung & lung\_small\_cell\_carcinoma & BMS-536924 & IGF1R & IGF1R \\
26 & MS-1 & lung & lung\_small\_cell\_carcinoma & BMS-509744 & ITK & ITK \\
27 & MZ7-mel & skin & melanoma & Pyrimethamine & Dihydrofolate reductase (DHFR) & DHFR \\
28 & NCI-H510A & lung & lung\_small\_cell\_carcinoma & JW-7-52-1 & MTOR & MTOR \\
29 & SK-MM-2 & blood & Myeloma & A-443654 & AKT1/2/3 & AKT \\
30 & TE-1 & aero\_digestive\_tract & oesophagus & GW843682X & PLK1 & PLK \\
31 & TE-10 & aero\_digestive\_tract & oesophagus & MS-275 & HDAC & HDAC \\
32 & HL-60 & blood & acute\_myeloid\_leukaemia & Parthenolide & NFKB1 & NFKB1 \\
33 & NCI-H226 & lung & lung\_NSCLC\_squamous\_cell\_carcinoma & KIN001-135 & IKKE & IKKE \\
34 & NCI-H23 & lung & lung\_NSCLC\_adenocarcinoma & TGX221 & PI3K beta & PI3K \\
35 & SR & blood & lymphoid\_neoplasm\_other & Bortezomib & Proteasome & Proteasome \\
36 & UACC-257 & skin & melanoma & XMD8-85 & ERK5 (MK07) & ERK5 \\
37 & TK10 & kidney & kidney & Roscovitine & CDKs & CDK \\
38 & SF268 & nervous\_system & glioma & Salubrinal & GADD34-PP1C phosphatase & GADD34-PP1C phosphatase \\
39 & KM12 & digestive\_system & large\_intestine & Lapatinib & EGFR, ERBB2 & EGFR \\
40 & Becker & nervous\_system & glioma & GSK269962A & ROCK & ROCK \\
41 & 697 & blood & lymphoblastic\_leukemia & Doxorubicin & DNA intercalating & DNA \\
42 & COR-L88 & lung & lung\_small\_cell\_carcinoma & Etoposide & TOP2 & TOP \\
43 & COLO-824 & breast & breast & Gemcitabine & DNA replication & DNA \\
44 & DG-75 & blood & Burkitt\_lymphoma & Mitomycin C & DNA crosslinker & DNA \\
45 & DJM-1 & skin & skin\_other & Vinorelbine & Microtubules & Microtubules \\
46 & DU-4475 & breast & breast & NSC-87877 & SHP1/2 (PTN6/11) & SHP1/2 (PTN6/11) \\
47 & EB2 & blood & Burkitt\_lymphoma & Bicalutamide & Androgen receptor (ANDR) & ANDR \\
48 & EB-3 & blood & Burkitt\_lymphoma & Midostaurin & KIT & PDGFR \\
49 & GCIY & digestive\_system & stomach & CHIR-99021 & GSK3B & GSK3B \\
50 & GI-ME-N & nervous\_system & neuroblastoma & AP-24534 & ABL & SRC,ABL \\
51 & GOTO & nervous\_system & neuroblastoma & AZD6482 & PI3Kb (P3C2B) & PI3K \\
52 & GT3TKB & digestive\_system & stomach & PF-562271 & FAK & FAK \\
53 & HEL & blood & acute\_myeloid\_leukaemia & DMOG & Prolyl-4-Hydroxylase & Hydroxylase \\
54 & HH & blood & T\_cell\_leukemia & FTI-277 & Farnesyl transferase (FNTA) & FNTA \\
55 & HT & blood & B\_cell\_lymphoma & OSU-03012 & PDK1 (PDPK1) & PDK1 \\
56 & IST-MES1 & lung & mesothelioma & Shikonin & unknown & unknown \\
57 & JiyoyeP-2003 & blood & Burkitt\_lymphoma & AKT inhibitor VIII & AKT1/2 & AKT \\
58 & JVM-2 & blood & lymphoid\_neoplasm\_other & Embelin & XIAP & XIAP \\
59 & KARPAS-299 & blood & anaplastic\_large\_cell\_lymphoma & FH535 & unknown & unknown \\
60 & KARPAS-422 & blood & B\_cell\_lymphoma & PAC-1 & CASP3 activator & CASP3 \\
61 & KINGS-1 & nervous\_system & glioma & IPA-3 & PAK & PAK \\
62 & KMOE-2 & blood & acute\_myeloid\_leukaemia & GSK-650394 & SGK3 & SGK3 \\
63 & KP-N-YN & nervous\_system & neuroblastoma & BAY 61-3606 & SYK & SYK \\
64 & L-428 & blood & Hodgkin\_lymphoma & Thapsigargin & ATPase, Ca++ transporting, cardiac muscle, slow twitch 2 & ATPase \\
65 & L-540 & blood & Hodgkin\_lymphoma & Obatoclax Mesylate & BCL-2, BCL-XL, MCL-1 & BCL \\
66 & LNCaP-Clone-FGC & urogenital\_system & prostate & BMS-754807 & IGF1R & IGF1R \\
67 & LS-123 & digestive\_system & large\_intestine & OSI-906 & IGF1R & IGF1R \\
68 & LS-411N & digestive\_system & large\_intestine & Bexarotene & Retinioic acid X family agonist & Retinoic \\
69 & LS-513 & digestive\_system & large\_intestine & Bleomycin & DNA damage & DNA \\
70 & MHH-PREB-1 & blood & B\_cell\_leukemia & LFM-A13 & BTK & BTK \\
71 & ML-2 & blood & acute\_myeloid\_leukaemia & AUY922 & HSP90 & HSP90 \\
72 & MN-60 & blood & B\_cell\_leukemia & Bryostatin 1 & PRKC & PRKC \\
73 & MONO-MAC-6 & blood & acute\_myeloid\_leukaemia & Pazopanib & VEGFR, PDGFRA, PDGFRB, KIT & PDGFR \\
74 & MPP-89 & lung & mesothelioma & LAQ824 & HDAC & HDAC \\
75 & NCI-SNU-5 & digestive\_system & stomach & Epothilone B & Microtubules & Microtubules \\
76 & NCI-SNU-16 & digestive\_system & stomach & GSK-1904529A & IGF1R & IGF1R \\
77 & NH-12 & nervous\_system & neuroblastoma & Tipifarnib & Farnesyl-transferase (FNTA) & FNTA \\
78 & NMC-G1 & nervous\_system & glioma & AS601245 & JNK & JNK \\
79 & no-11 & nervous\_system & glioma & AICAR & AMPK agonist & AMPK \\
80 & NOMO-1 & blood & acute\_myeloid\_leukaemia & Camptothecin & TOP1 & TOP \\
81 & NCI-H524 & lung & lung\_small\_cell\_carcinoma & Vinblastine & Microtubules & Microtubules \\
82 & P30-OHK & blood & lymphoblastic\_leukemia & Cisplatin & DNA crosslinker & DNA \\
83 & P31-FUJ & blood & acute\_myeloid\_leukaemia & Cytarabine & DNA synthesis & DNA \\
84 & PF-382 & blood & lymphoblastic\_T\_cell\_leukaemia & Docetaxel & Microtubules & Microtubules \\
85 & Raji & blood & Burkitt\_lymphoma & Methotrexate & Dihydrofolate reductase (DHFR) & DHFR \\
86 & REH & blood & B\_cell\_leukemia & ATRA & Retinoic acid and retinoid X receptor agonist & Retinoic \\
87 & SF126 & nervous\_system & glioma & Gefitinib & EGFR & EGFR \\
88 & SJSA-1 & bone & bone\_other & ABT-263 & BCL2, BCL-XL, BCL-W & BCL \\
89 & SK-NEP-1 & kidney & kidney & Vorinostat & HDAC inhibitor Class I, IIa, IIb, IV & HDAC \\
90 & SW684 & soft\_tissue & fibrosarcoma & Nilotinib & ABL & SRC,ABL \\
91 & SW872 & soft\_tissue & soft\_tissue\_other & RDEA119 & MEK1/2 & MEK \\
92 & SW962 & urogenital\_system & urogenital\_system\_other & CI-1040 & MEK1/2 & MEK \\
93 & TUR & blood & B\_cell\_lymphoma & Temsirolimus & MTOR & MTOR \\
94 & U-698-M & blood & B\_cell\_leukemia & AZD-2281 & PARP1/2 & PARP \\
95 & WSU-NHL & blood & lymphoid\_neoplasm\_other & ABT-888 & PARP1/2 & PARP \\
96 & RCC10RGB & kidney & kidney & Bosutinib & SRC, ABL, TEC & SRC,ABL \\
97 & KURAMOCHI & urogenital\_system & ovary & Lenalidomide & TNF alpha & TNF \\
98 & Ramos-2G6-4C10 & blood & Burkitt\_lymphoma & Axitinib & PDGFR, KIT, VEGFR & PDGFR \\
99 & CW-2 & digestive\_system & large\_intestine & CEP-701 & FLT3, JAK2, NTRK1, RET & FLT3 \\
100 & COLO-320-HSR & digestive\_system & large\_intestine & 17-AAG & HSP90 & HSP90 \\
101 & COLO-684 & urogenital\_system & endometrium & VX-702 & p38 & JNK \\
102 & CA46 & blood & Burkitt\_lymphoma & AMG-706 & VEGFR, RET, c-KIT, PDGFR & PDGFR \\
103 & RL & blood & B\_cell\_lymphoma & KU-55933 & ATM & ATM \\
104 & SNU-C1 & digestive\_system & large\_intestine & BIBW2992 & EGFR, ERBB2 & EGFR \\
105 & ST486 & blood & Burkitt\_lymphoma & GDC-0449 & SMO & SMO \\
106 & BC-1 & blood & B\_cell\_lymphoma & PLX4720 & BRAF & BRAF \\
107 & CAS-1 & nervous\_system & glioma & BX-795 & TBK1, PDK1, IKK, AURKB/C & PDK1 \\
108 & MFM-223 & breast & breast & NU-7441 & DNAPK & DNAPK \\
109 & LS-1034 & digestive\_system & large\_intestine & BIRB 0796 & p38, JNK2 & JNK \\
110 & HDLM-2 & blood & Hodgkin\_lymphoma & JNK Inhibitor VIII & JNK & JNK \\
111 & KNS-81-FD & nervous\_system & glioma & 681640 & WEE1, CHK1 & CHK1 \\
112 & TE-6 & aero\_digestive\_tract & oesophagus & Nutlin-3a & MDM2 & MDM \\
113 & TE-12 & aero\_digestive\_tract & oesophagus & PD-173074 & FGFR1/3 & FGFR \\
114 & D-263MG & nervous\_system & glioma & ZM-447439 & AURKB & PDK1 \\
115 & D-502MG & nervous\_system & glioma & RO-3306 & CDK1 & CDK \\
116 & EW-3 & bone & ewings\_sarcoma & PD-0332991 & CDK4/6 & CDK \\
117 & EW-1 & bone & ewings\_sarcoma & GDC0941 & PI3K (class 1) & PI3K \\
118 & EW-18 & bone & ewings\_sarcoma & PD-0325901 & MEK1/2 & MEK \\
119 & EW-24 & bone & ewings\_sarcoma & SB590885 & BRAF & BRAF \\
120 & LAN-6 & nervous\_system & neuroblastoma & AZD6244 & MEK1/2 & MEK \\
121 & NB10 & nervous\_system & neuroblastoma & BMS-708163 & gamma-secretase & G-secretase \\
122 & NB6 & nervous\_system & neuroblastoma & JNJ-26854165 & MDM2 & MDM \\
123 & NB5 & nervous\_system & neuroblastoma & TW 37 & BCL-2, BCL-XL & BCL \\
124 & NB14 & nervous\_system & neuroblastoma & AG-014699 & PARP1, PARP2 & PARP \\ 
\hline
\end{tabular}%
}
\caption{Cell lines and drugs used in the GDSC case study.}
\label{SI-tab-GDSC-Dataset}
\end{table}
\begin{table}[]
\centering
\begin{adjustbox}{width=1\textwidth,height=0.30\textheight}
\begin{tabular}{@{}lll@{}}
\hline
 & Pathway Name & Reference \\ 
 \hline
1 & BIOCARTA\_EGFR\_SMRTE\_PATHWAY & http://www.broadinstitute.org/gsea/msigdb/cards/BIOCARTA\_EGFR\_SMRTE\_PATHWAY \\
2 & PID\_VEGF\_VEGFR\_PATHWAY & http://www.broadinstitute.org/gsea/msigdb/cards/PID\_VEGF\_VEGFR\_PATHWAY \\
3 & REACTOME\_SIGNALING\_BY\_EGFR\_IN\_CANCER & http://www.broadinstitute.org/gsea/msigdb/cards/REACTOME\_SIGNALING\_BY\_EGFR\_IN\_CANCER \\
4 & REACTOME\_EGFR\_DOWNREGULATION & http://www.broadinstitute.org/gsea/msigdb/cards/REACTOME\_EGFR\_DOWNREGULATION \\
5 & REACTOME\_SHC1\_EVENTS\_IN\_EGFR\_SIGNALING & http://www.broadinstitute.org/gsea/msigdb/cards/REACTOME\_SHC1\_EVENTS\_IN\_EGFR\_SIGNALING \\
6 & KEGG\_MTOR\_SIGNALING\_PATHWAY & http://www.broadinstitute.org/gsea/msigdb/cards/KEGG\_MTOR\_SIGNALING\_PATHWAY \\
7 & BIOCARTA\_MTOR\_PATHWAY & http://www.broadinstitute.org/gsea/msigdb/cards/BIOCARTA\_MTOR\_PATHWAY \\
8 & BIOCARTA\_IGF1MTOR\_PATHWAY & http://www.broadinstitute.org/gsea/msigdb/cards/BIOCARTA\_IGF1MTOR\_PATHWAY \\
9 & PID\_MTOR\_4PATHWAY & http://www.broadinstitute.org/gsea/msigdb/cards/PID\_MTOR\_4PATHWAY \\
10 & REACTOME\_ENERGY\_DEPENDENT\_REGULATION\_OF\_MTOR\_BY\_LKB1\_AMPK & http://www.broadinstitute.org/gsea/msigdb/cards/REACTOME\_ENERGY\_DEPENDENT\_REGULATION\_OF\_MTOR\_BY\_LKB1\_AMPK \\
11 & PID\_KIT\_PATHWAY & http://www.broadinstitute.org/gsea/msigdb/cards/PID\_KIT\_PATHWAY \\
12 & REACTOME\_REGULATION\_OF\_KIT\_SIGNALING & http://www.broadinstitute.org/gsea/msigdb/cards/REACTOME\_REGULATION\_OF\_KIT\_SIGNALING \\
13 & BIOCARTA\_MET\_PATHWAY & http://www.broadinstitute.org/gsea/msigdb/cards/BIOCARTA\_MET\_PATHWAY \\
14 & PID\_MET\_PATHWAY & http://www.broadinstitute.org/gsea/msigdb/cards/PID\_MET\_PATHWAY \\
15 & BIOCARTA\_AKT\_PATHWAY & http://www.broadinstitute.org/gsea/msigdb/cards/BIOCARTA\_AKT\_PATHWAY \\
16 & PID\_PI3KCI\_AKT\_PATHWAY & http://www.broadinstitute.org/gsea/msigdb/cards/PID\_PI3KCI\_AKT\_PATHWAY \\
17 & REACTOME\_NEGATIVE\_REGULATION\_OF\_THE\_PI3K\_AKT\_NETWORK & http://www.broadinstitute.org/gsea/msigdb/cards/REACTOME\_NEGATIVE\_REGULATION\_OF\_THE\_PI3K\_AKT\_NETWORK \\
18 & REACTOME\_PI3K\_AKT\_ACTIVATION & http://www.broadinstitute.org/gsea/msigdb/cards/REACTOME\_PI3K\_AKT\_ACTIVATION \\
19 & REACTOME\_AKT\_PHOSPHORYLATES\_TARGETS\_IN\_THE\_CYTOSOL & http://www.broadinstitute.org/gsea/msigdb/cards/REACTOME\_AKT\_PHOSPHORYLATES\_TARGETS\_IN\_THE\_CYTOSOL \\
20 & REACTOME\_CD28\_DEPENDENT\_PI3K\_AKT\_SIGNALING & http://www.broadinstitute.org/gsea/msigdb/cards/REACTOME\_CD28\_DEPENDENT\_PI3K\_AKT\_SIGNALING \\
21 & REACTOME\_PIP3\_ACTIVATES\_AKT\_SIGNALING & http://www.broadinstitute.org/gsea/msigdb/cards/REACTOME\_PIP3\_ACTIVATES\_AKT\_SIGNALING \\
22 & BIOCARTA\_HDAC\_PATHWAY & http://www.broadinstitute.org/gsea/msigdb/cards/BIOCARTA\_HDAC\_PATHWAY \\
23 & PID\_HDAC\_CLASSII\_PATHWAY & http://www.broadinstitute.org/gsea/msigdb/cards/PID\_HDAC\_CLASSII\_PATHWAY \\
24 & PID\_HDAC\_CLASSIII\_PATHWAY & http://www.broadinstitute.org/gsea/msigdb/cards/PID\_HDAC\_CLASSIII\_PATHWAY \\
25 & PID\_HDAC\_CLASSI\_PATHWAY & http://www.broadinstitute.org/gsea/msigdb/cards/PID\_HDAC\_CLASSI\_PATHWAY \\
26 & PID\_ERBB2\_ERBB3\_PATHWAY & http://www.broadinstitute.org/gsea/msigdb/cards/PID\_ERBB2\_ERBB3\_PATHWAY \\
27 & REACTOME\_DOWNREGULATION\_OF\_ERBB2\_ERBB3\_SIGNALING & http://www.broadinstitute.org/gsea/msigdb/cards/REACTOME\_DOWNREGULATION\_OF\_ERBB2\_ERBB3\_SIGNALING \\
28 & REACTOME\_GRB2\_EVENTS\_IN\_ERBB2\_SIGNALING & http://www.broadinstitute.org/gsea/msigdb/cards/REACTOME\_GRB2\_EVENTS\_IN\_ERBB2\_SIGNALING \\
29 & REACTOME\_PI3K\_EVENTS\_IN\_ERBB2\_SIGNALING & http://www.broadinstitute.org/gsea/msigdb/cards/REACTOME\_PI3K\_EVENTS\_IN\_ERBB2\_SIGNALING \\
30 & ST\_JNK\_MAPK\_PATHWAY & http://www.broadinstitute.org/gsea/msigdb/cards/ST\_JNK\_MAPK\_PATHWAY \\
31 & PID\_TCR\_JNK\_PATHWAY & http://www.broadinstitute.org/gsea/msigdb/cards/PID\_TCR\_JNK\_PATHWAY \\
32 & REACTOME\_JNK\_C\_JUN\_KINASES\_PHOSPHORYLATION\_AND\_ACTIVATION\_MEDIATED\_BY\_ACTIVATED\_HUMAN\_TAK1 & http://www.broadinstitute.org/gsea/msigdb/cards/REACTOME\_JNK\_C\_JUN\_KINASES\_PHOSPHORYLATION\_AND\_ACTIVATION\_MEDIATED\_BY\_ACTIVATED\_HUMAN\_TAK1 \\
33 & BIOCARTA\_ATM\_PATHWAY & http://www.broadinstitute.org/gsea/msigdb/cards/BIOCARTA\_ATM\_PATHWAY \\
34 & PID\_ATM\_PATHWAY & http://www.broadinstitute.org/gsea/msigdb/cards/PID\_ATM\_PATHWAY \\
35 & REACTOME\_IRAK1\_RECRUITS\_IKK\_COMPLEX & http://www.broadinstitute.org/gsea/msigdb/cards/REACTOME\_IRAK1\_RECRUITS\_IKK\_COMPLEX \\
36 & REACTOME\_ACTIVATION\_OF\_IRF3\_IRF7\_MEDIATED\_BY\_TBK1\_IKK\_EPSILON & http://www.broadinstitute.org/gsea/msigdb/cards/REACTOME\_ACTIVATION\_OF\_IRF3\_IRF7\_MEDIATED\_BY\_TBK1\_IKK\_EPSILON \\
37 & REACTOME\_IKK\_COMPLEX\_RECRUITMENT\_MEDIATED\_BY\_RIP1 & http://www.broadinstitute.org/gsea/msigdb/cards/REACTOME\_IKK\_COMPLEX\_RECRUITMENT\_MEDIATED\_BY\_RIP1 \\
38 & REACTOME\_NEGATIVE\_REGULATION\_OF\_FGFR\_SIGNALING & http://www.broadinstitute.org/gsea/msigdb/cards/REACTOME\_NEGATIVE\_REGULATION\_OF\_FGFR\_SIGNALING \\
39 & REACTOME\_SIGNALING\_BY\_FGFR\_IN\_DISEASE & http://www.broadinstitute.org/gsea/msigdb/cards/REACTOME\_SIGNALING\_BY\_FGFR\_IN\_DISEASE \\
40 & REACTOME\_SIGNALING\_BY\_FGFR\_MUTANTS & http://www.broadinstitute.org/gsea/msigdb/cards/REACTOME\_SIGNALING\_BY\_FGFR\_MUTANTS \\
41 & REACTOME\_FGFR\_LIGAND\_BINDING\_AND\_ACTIVATION & http://www.broadinstitute.org/gsea/msigdb/cards/REACTOME\_FGFR\_LIGAND\_BINDING\_AND\_ACTIVATION \\
42 & PID\_IL2\_PI3K\_PATHWAY & http://www.broadinstitute.org/gsea/msigdb/cards/PID\_IL2\_PI3K\_PATHWAY \\
43 & PID\_PI3K\_PLC\_TRK\_PATHWAY & http://www.broadinstitute.org/gsea/msigdb/cards/PID\_PI3K\_PLC\_TRK\_PATHWAY \\
44 & REACTOME\_PI3K\_EVENTS\_IN\_ERBB4\_SIGNALING & http://www.broadinstitute.org/gsea/msigdb/cards/REACTOME\_PI3K\_EVENTS\_IN\_ERBB4\_SIGNALING \\
45 & REACTOME\_PI3K\_CASCADE & http://www.broadinstitute.org/gsea/msigdb/cards/REACTOME\_PI3K\_CASCADE \\
46 & EGFR\_UP.V1\_DN & http://www.broadinstitute.org/gsea/msigdb/cards/EGFR\_UP.V1\_DN \\
47 & EGFR\_UP.V1\_UP & http://www.broadinstitute.org/gsea/msigdb/cards/EGFR\_UP.V1\_UP \\
48 & AKT\_UP\_MTOR\_DN.V1\_DN & http://www.broadinstitute.org/gsea/msigdb/cards/AKT\_UP\_MTOR\_DN.V1\_DN \\
49 & AKT\_UP\_MTOR\_DN.V1\_UP & http://www.broadinstitute.org/gsea/msigdb/cards/AKT\_UP\_MTOR\_DN.V1\_UP \\
50 & MTOR\_UP.V1\_DN & http://www.broadinstitute.org/gsea/msigdb/cards/MTOR\_UP.V1\_DN \\
51 & MTOR\_UP.V1\_UP & http://www.broadinstitute.org/gsea/msigdb/cards/MTOR\_UP.V1\_UP \\
52 & MTOR\_UP.N4.V1\_DN & http://www.broadinstitute.org/gsea/msigdb/cards/MTOR\_UP.N4.V1\_DN \\
53 & MTOR\_UP.N4.V1\_UP & http://www.broadinstitute.org/gsea/msigdb/cards/MTOR\_UP.N4.V1\_UP \\
54 & JAK2\_DN.V1\_DN & http://www.broadinstitute.org/gsea/msigdb/cards/JAK2\_DN.V1\_DN \\
55 & JAK2\_DN.V1\_UP & http://www.broadinstitute.org/gsea/msigdb/cards/JAK2\_DN.V1\_UP \\
56 & ALK\_DN.V1\_DN & http://www.broadinstitute.org/gsea/msigdb/cards/ALK\_DN.V1\_DN \\
57 & ALK\_DN.V1\_UP & http://www.broadinstitute.org/gsea/msigdb/cards/ALK\_DN.V1\_UP \\
58 & SRC\_UP.V1\_DN & http://www.broadinstitute.org/gsea/msigdb/cards/SRC\_UP.V1\_DN \\
59 & SRC\_UP.V1\_UP & http://www.broadinstitute.org/gsea/msigdb/cards/SRC\_UP.V1\_UP \\
60 & AKT\_UP.V1\_DN & http://www.broadinstitute.org/gsea/msigdb/cards/AKT\_UP.V1\_DN \\
61 & AKT\_UP.V1\_UP & http://www.broadinstitute.org/gsea/msigdb/cards/AKT\_UP.V1\_UP \\
62 & JNK\_DN.V1\_DN & http://www.broadinstitute.org/gsea/msigdb/cards/JNK\_DN.V1\_DN \\
63 & JNK\_DN.V1\_UP & http://www.broadinstitute.org/gsea/msigdb/cards/JNK\_DN.V1\_UP \\
64 & MEK\_UP.V1\_DN & http://www.broadinstitute.org/gsea/msigdb/cards/MEK\_UP.V1\_DN \\
65 & MEK\_UP.V1\_UP & http://www.broadinstitute.org/gsea/msigdb/cards/MEK\_UP.V1\_UP \\
66 & ATM\_DN.V1\_DN & http://www.broadinstitute.org/gsea/msigdb/cards/ATM\_DN.V1\_DN \\
67 & ATM\_DN.V1\_UP & http://www.broadinstitute.org/gsea/msigdb/cards/ATM\_DN.V1\_UP \\
68 & TBK1.DF\_DN & http://www.broadinstitute.org/gsea/msigdb/cards/TBK1.DF\_DN \\
69 & TBK1.DF\_UP & http://www.broadinstitute.org/gsea/msigdb/cards/TBK1.DF\_UP \\
70 & TBK1.DN.48HRS\_DN & http://www.broadinstitute.org/gsea/msigdb/cards/TBK1.DN.48HRS\_DN \\
71 & TBK1.DN.48HRS\_UP & http://www.broadinstitute.org/gsea/msigdb/cards/TBK1.DN.48HRS\_UP \\ 
\hline
\end{tabular}%
\end{adjustbox}
\caption{List of pathways used in the GDSC case study. Some of the names are modified so that they can fit into the eye diagram (main text Figure 5). For example, reactome SHC1 events in EGFR signaling is modified as re:SHC1 event events in EGFR signaling.}
\label{SI-tab-GDSC-Pathways}
\end{table}

\begin{table}[h]
	\caption{Statistical significance of the predictive performances on the GDSC data set. P-values from one-sided paired Wilcoxon Sign-Rank test corresponding to the curves shown in Figure 2 (in the main text), corrected for multiple testing using Benjamini, Hochberg, and Yekutieli's method~\cite{benjamini1995controlling},\cite{benjamini2001control}.}\label{tab:pvalues-sanger_cv_pred}
	\vspace{5mm}
		\begin{tabular}{lcc}
	
				Methods & {\it cw}KBMF$_{single-view}$ & {\it cw}KBMF$_{multi-view}$ \\
				\hline
				{\it cw}KBMF$_{single-view}$ & -- & $1.369 \times 10^{-6}$ \\
				KBMF$_{multi-view}$ & -- & $1.369 \times 10^{-6}$ \\
				BMTMKL$_{multi-view}$ & -- & $2.922 \times 10^{-2}$ \\
				KPMF$_{single-view}$ & $1.369 \times 10^{-6}$ & $1.369 \times 10^{-6}$ \\
				MT-LR$_{single-view}$ & $2.017\times 10^{-4}$ & $2.790 \times 10^{-5}$ \\
				Baseline & $1.369 \times 10^{-6}$ & $1.369 \times 10^{-6}$ \\
				\hline
			\end{tabular}
\end{table}


\clearpage
\subsection*{Case Study: CTRP Data Set}
\begin{figure*}[ht]
\begin{center}
	\includegraphics[width=1\textwidth,height=0.5\textheight,keepaspectratio]{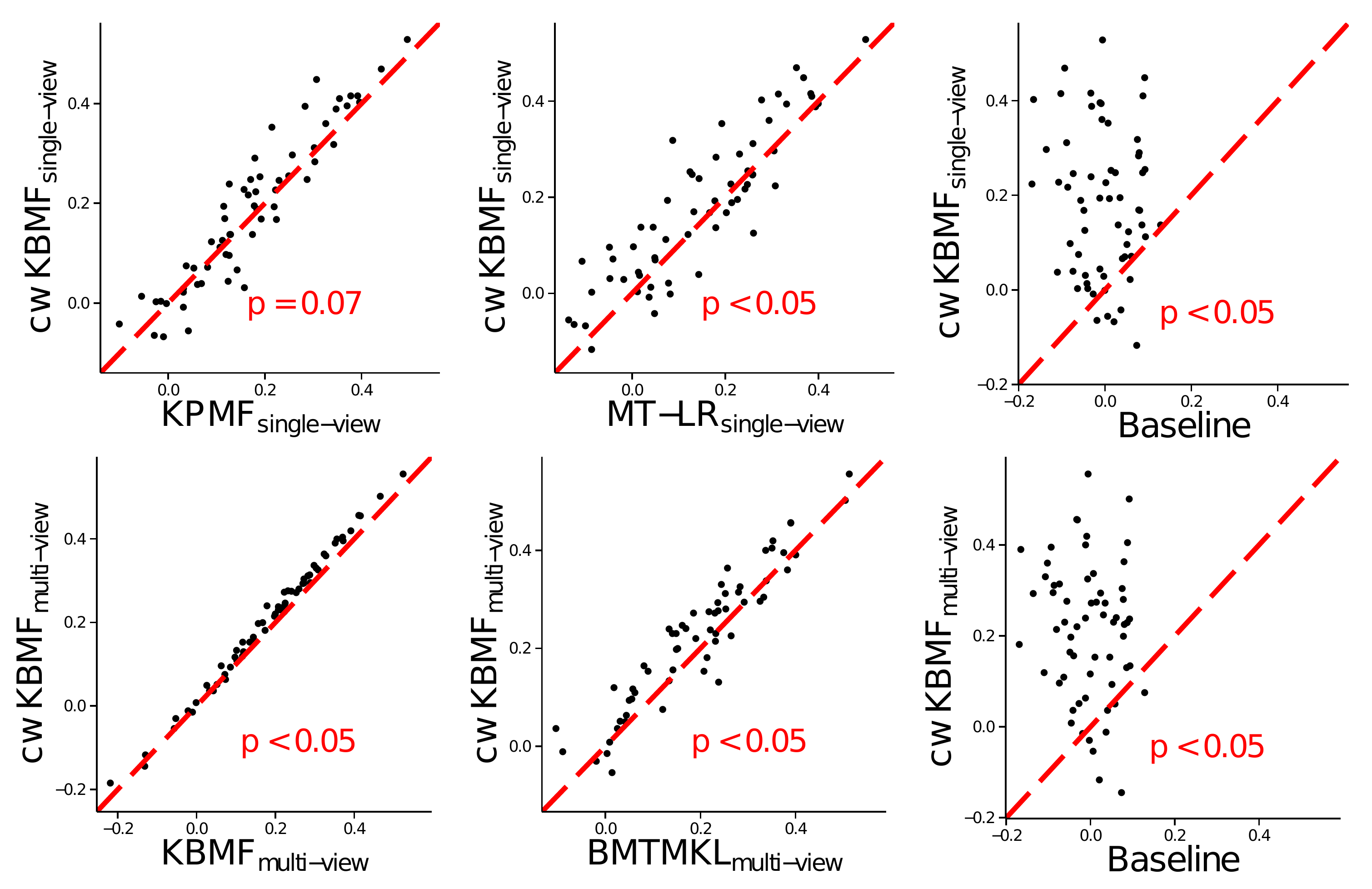}
\caption{Comparison of {\it cw}KBMF with other methods on CTRP data set. Spearman rank correlation between the predictions and the response data, for each drug are calculated for each method. The correlations from each method (x-axis) are compared to those from {\it cw}KBMF (y-axis) with components 10. Each dot represents cross validated prediction performance for a drug response averaged over 10 random rounds of 5 fold cross-validation procedure. The subscript in method's name denotes the format of learning data used by the methods; {\it single-view} means that the method is learned using one view (i.e, one group containing all the genomic features), while {\it multi-view} means that the method is learned using multiple views (i.e., genomic features are grouped into many views based on the prior knowledge about the pathways).
	Method abbreviation: {\it cw}KBMF, kernelized Bayesian matrix factorization with component-wise MKL; KBMF, kernelized Bayesian matrix factorization; BMTMKL, Bayesian multi-task MKL; KPMF, kernelized probabilistic matrix factorization; MT-LR, multi-task sparse linear regression; Baseline, mean of the training data. P-values show the statistical significance of the improvements of {\it cw}KBMF’s predictions compared to other methods computed with one-sided paired Wilcoxon Sign-Rank test.}
	\label{fig:ctrp_pred_k}
\end{center}
\end{figure*}

Below, we provides supplementary information of the data and results from the CTRP data set as discussed in the main text section 3 and 4.\\
Table~\ref{SI-tab-CTRP-Dataset} shows the names, types and tissues of 66 cancer cell lines and names, primary targets of 63 drugs used in our study. We obtained these data from Cancer Therapeutic Response Portal~\cite{basu2013interactive}, and show here for the purpose of motivating reproducibility of the results. Table~\ref{SI-tab-CTRP-Pathways} list the 25 pathways used in the case study.\\
Table~\ref{tab:pvalues-ctrp_cv_pred} shows the statistical significance of the predictive performances of {\it cw}KBMF compared to other methods on the GDSC data set. Figure~\ref{fig:ctrp_pred_k} compares the predictive performance of {\it cw}KBMF with other methods on CTRP data, when learned with 20 components. 

\begin{table}[h]
\centering
\resizebox{\textwidth}{!}{%
\begin{tabular}{@{}llllll@{}}
\toprule
 & Cell line & Type & Tissue & Drug & Primary Target \\ \midrule
1 & A2780 & OVARY &  & etoposide & TOP2A;TOP2B \\
2 & A375 & SKIN &  & oligomycin A & ATP5L2 \\
3 & AMO1 & HEMATOPOIETIC\_AND\_LYMPHOID\_TISSUE & PLASMA\_CELL\_MYELOMA & chaetocin & SUV39H1 \\
4 & BL70 & HEMATOPOIETIC\_AND\_LYMPHOID\_TISSUE &  & tubastatin A & HDAC6 \\
5 & CAL12T & LUNG & NON\_SMALL\_CELL\_CARCINOMA & avrainvillamide & NPM1 \\
6 & COLO320 & LARGE\_INTESTINE &  & TG-101348 & JAK2 \\
7 & COLO678 & LARGE\_INTESTINE &  & SCH-79797 & F2R \\
8 & CORL23 & LUNG & LARGE\_CELL\_CARCINOMA & HLI 373 & MDM2 \\
9 & COV318 & OVARY &  & indisulam & CA9 \\
10 & EFO27 & OVARY &  & Compound 10b & MAP3K8 \\
11 & GAMG & CENTRAL\_NERVOUS\_SYSTEM &  & BMS-536924 & IGF1R;INSR \\
12 & HCC15 & LUNG & SQUAMOUS\_CELL\_CARCINOMA & Compound 12 & NOS2 \\
13 & HCC2935 & LUNG & NON\_SMALL\_CELL\_CARCINOMA & JZL-184 & MGLL \\
14 & HCC4006 & LUNG & ADENOCARCINOMA & cerulenin & FASN;HMGCS1 \\
15 & HCC827 & LUNG & ADENOCARCINOMA & pemetrexed & TS;DHFR;GART \\
16 & HCT116 & LARGE\_INTESTINE &  & serdemetan & MDM2 \\
17 & HEC151 & ENDOMETRIUM &  & tipifarnib-P1 & FNTA \\
18 & HEC59 & ENDOMETRIUM &  & ATRA & RARA;RARB;RARG \\
19 & HEPG2 & LIVER &  & SNX-2112 & HSP90AA1;HSP90B1 \\
20 & HS852T & SKIN &  & BRD4770 & EHMT2 \\
21 & HT1080 & SOFT\_TISSUE &  & NSC632839 & USP5;USP13 \\
22 & HT29 & LARGE\_INTESTINE &  & WT-161 & HDAC6 \\
23 & IGROV1 & OVARY &  & losartan & AGTR2 \\
24 & JHH6 & LIVER &  & vorinostat & HDAC1;HDAC2;HDAC3;HDAC6;HDAC8 \\
25 & JHOS4 & OVARY &  & batimastat & MMP1;MMP2;MMP3;MMP7;MMP9 \\
26 & JHUEM2 & ENDOMETRIUM &  & PF-750 & FAAH \\
27 & KLE & ENDOMETRIUM &  & 2-bromopyruvate & HK2 \\
28 & KMS11 & HEMATOPOIETIC\_AND\_LYMPHOID\_TISSUE & PLASMA\_CELL\_MYELOMA & AC55649 & RARB \\
29 & LN229 & CENTRAL\_NERVOUS\_SYSTEM &  & necrostatin-1 & RIPK1 \\
30 & LP1 & HEMATOPOIETIC\_AND\_LYMPHOID\_TISSUE & PLASMA\_CELL\_MYELOMA & PX-12 & TXN \\
31 & LS123 & LARGE\_INTESTINE &  & nimodipine & CACNA1C;CACNA1D;CACNA1S;CACNA1F \\
32 & MCF7 & BREAST &  & lovastatin acid & HMGCR \\
33 & MKN74 & STOMACH &  & GW-405833 & CB2 \\
34 & NCIH1694 & LUNG & SMALL\_CELL\_CARCINOMA & minoxidil & KCNJ8;KCNJ11 \\
35 & NCIH1915 & LUNG & LARGE\_CELL\_CARCINOMA & PRIMA-1 & TP53 \\
36 & NCIH2023 & LUNG & ADENOCARCINOMA & CR-1-31B & EIF4A2;EIF4E;EIF4G1 \\
37 & NCIH2030 & LUNG & NON\_SMALL\_CELL\_CARCINOMA & CHEMBL399379 & ACLY \\
38 & NCIH2122 & LUNG & ADENOCARCINOMA & sildenafil & PDE5A \\
39 & NCIH2172 & LUNG & NON\_SMALL\_CELL\_CARCINOMA & selumetinib & MAP2K1;MAP2K2 \\
40 & NCIH2286 & LUNG & SMALL\_CELL\_CARCINOMA & TGX-115 & PIK3CB;PIK3CD \\
41 & NCIH3255 & LUNG & ADENOCARCINOMA & 10-DEBC & AKT \\
42 & NCIH441 & LUNG & ADENOCARCINOMA & BML-259 & CDK5;CDK2 \\
43 & NCIH460 & LUNG & LARGE\_CELL\_CARCINOMA & PF-04217903 & MET \\
44 & OCILY10 & HEMATOPOIETIC\_AND\_LYMPHOID\_TISSUE & DIFFUSE\_LARGE\_B\_CELL\_LYMPHOMA & purmorphamine & SMO \\
45 & OV90 & OVARY &  & RG-108 & DNMT1 \\
46 & OVCAR4 & OVARY &  & tosedostat & LAP3;NPEPPS;ANPEP \\
47 & OVCAR8 & OVARY &  & eflornithine & ODC1 \\
48 & OVMANA & OVARY &  & LY-2365109 & SLC6A9 \\
49 & PATU8902 & PANCREAS &  & ML090 & NOX1 \\
50 & PC3 & PROSTATE &  & bovinocidin & SDHA;SDHB;SDHC;SDHD \\
51 & RKN & SOFT\_TISSUE &  & irosustat & STS \\
52 & RKO & LARGE\_INTESTINE &  & Compound 2 & BCAT1 \\
53 & SCC9 & UPPER\_AERODIGESTIVE\_TRACT &  & importazole & KPNB1 \\
54 & SKLU1 & LUNG & ADENOCARCINOMA & YK 4-279 & DHX9;ERG;ETV1 \\
55 & SKMEL5 & SKIN &  & AG14361 & PARP1 \\
56 & SNGM & ENDOMETRIUM &  & narciclasine & RHOA \\
57 & SNU398 & LIVER &  & 3-aminobenzamide & PARP1 \\
58 & SNU449 & LIVER &  & B02 & RAD51 \\
59 & SQ1 & LUNG & SQUAMOUS\_CELL\_CARCINOMA & PJ 34 & PARP1 \\
60 & SUDHL4 & HEMATOPOIETIC\_AND\_LYMPHOID\_TISSUE & DIFFUSE\_LARGE\_B\_CELL\_LYMPHOMA & BRD8899 & STK33 \\
61 & SW480 & LARGE\_INTESTINE &  & MK-2206 & AKT \\
62 & SW620 & LARGE\_INTESTINE &  & MLN2238 & PSMB5 \\
63 & T3M10 & LUNG & LARGE\_CELL\_CARCINOMA & CAY10618 & NAMPT \\
64 & TYKNU & OVARY &  & NA & NA \\
65 & U2OS & BONE &  & NA & NA \\
66 & U937 & HEMATOPOIETIC\_AND\_LYMPHOID\_TISSUE & DIFFUSE\_LARGE\_B\_CELL\_LYMPHOMA & NA & NA \\ \bottomrule
\end{tabular}%
}
\caption{Cell lines and drugs used in the CTRP case study.}
\label{SI-tab-CTRP-Dataset}
\end{table}
\begin{table}[]
\centering
\begin{adjustbox}{width=1\textwidth}
\begin{tabular}{@{}lll@{}}
\toprule
 & Pathway Name & Reference \\ \midrule
1 & BIOCARTA\_AKT\_PATHWAY & http://www.broadinstitute.org/gsea/msigdb/cards/BIOCARTA\_AKT\_PATHWAY \\
2 & PID\_PI3KCI\_AKT\_PATHWAY & http://www.broadinstitute.org/gsea/msigdb/cards/PID\_PI3KCI\_AKT\_PATHWAY \\
3 & REACTOME\_NEGATIVE\_REGULATION\_OF\_THE\_PI3K\_AKT\_NETWORK & http://www.broadinstitute.org/gsea/msigdb/cards/REACTOME\_NEGATIVE\_REGULATION\_OF\_THE\_PI3K\_AKT\_NETWORK \\
4 & REACTOME\_PI3K\_AKT\_ACTIVATION & http://www.broadinstitute.org/gsea/msigdb/cards/REACTOME\_PI3K\_AKT\_ACTIVATION \\
5 & REACTOME\_AKT\_PHOSPHORYLATES\_TARGETS\_IN\_THE\_CYTOSOL & http://www.broadinstitute.org/gsea/msigdb/cards/REACTOME\_AKT\_PHOSPHORYLATES\_TARGETS\_IN\_THE\_CYTOSOL \\
6 & REACTOME\_CD28\_DEPENDENT\_PI3K\_AKT\_SIGNALING & http://www.broadinstitute.org/gsea/msigdb/cards/REACTOME\_CD28\_DEPENDENT\_PI3K\_AKT\_SIGNALING \\
7 & REACTOME\_PIP3\_ACTIVATES\_AKT\_SIGNALING & http://www.broadinstitute.org/gsea/msigdb/cards/REACTOME\_PIP3\_ACTIVATES\_AKT\_SIGNALING \\
8 & BIOCARTA\_MET\_PATHWAY & http://www.broadinstitute.org/gsea/msigdb/cards/BIOCARTA\_MET\_PATHWAY \\
9 & PID\_MET\_PATHWAY & http://www.broadinstitute.org/gsea/msigdb/cards/PID\_MET\_PATHWAY \\
10 & PID\_RHOA\_PATHWAY & http://www.broadinstitute.org/gsea/msigdb/cards/PID\_RHOA\_PATHWAY \\
11 & PID\_RHOA\_REG\_PATHWAY & http://www.broadinstitute.org/gsea/msigdb/cards/PID\_RHOA\_REG\_PATHWAY \\
12 & JAK2\_DN.V1\_DN & http://www.broadinstitute.org/gsea/msigdb/cards/JAK2\_DN.V1\_DN \\
13 & JAK2\_DN.V1\_UP & http://www.broadinstitute.org/gsea/msigdb/cards/JAK2\_DN.V1\_UP \\
14 & EIF4E\_DN & http://www.broadinstitute.org/gsea/msigdb/cards/EIF4E\_DN \\
15 & EIF4E\_UP & http://www.broadinstitute.org/gsea/msigdb/cards/EIF4E\_UP \\
16 & AKT\_UP\_MTOR\_DN.V1\_DN & http://www.broadinstitute.org/gsea/msigdb/cards/AKT\_UP\_MTOR\_DN.V1\_DN \\
17 & AKT\_UP\_MTOR\_DN.V1\_UP & http://www.broadinstitute.org/gsea/msigdb/cards/AKT\_UP\_MTOR\_DN.V1\_UP \\
18 & AKT\_UP.V1\_DN & http://www.broadinstitute.org/gsea/msigdb/cards/AKT\_UP.V1\_DN \\
19 & AKT\_UP.V1\_UP & http://www.broadinstitute.org/gsea/msigdb/cards/AKT\_UP.V1\_UP \\
20 & STK33\_DN & http://www.broadinstitute.org/gsea/msigdb/cards/STK33\_DN \\
21 & STK33\_NOMO\_DN & http://www.broadinstitute.org/gsea/msigdb/cards/STK33\_NOMO\_DN \\
22 & STK33\_NOMO\_UP & http://www.broadinstitute.org/gsea/msigdb/cards/STK33\_NOMO\_UP \\
23 & STK33\_SKM\_DN & http://www.broadinstitute.org/gsea/msigdb/cards/STK33\_SKM\_DN \\
24 & STK33\_SKM\_UP & http://www.broadinstitute.org/gsea/msigdb/cards/STK33\_SKM\_UP \\
25 & STK33\_UP & http://www.broadinstitute.org/gsea/msigdb/cards/STK33\_UP \\ \bottomrule
\end{tabular}%
\end{adjustbox}
\caption{List of pathways used in the CTRP case study}
\label{SI-tab-CTRP-Pathways}
\end{table}
\begin{table}[h]
	\caption{Statistical significance of the predictive performances on the CTRP data set. P-values from one-sided paired Wilcoxon Sign-Rank test corresponding to the curves shown in Figure 2 in the main text, corrected for multiple testing using Benjamini, Hochberg, and Yekutieli's method~\cite{benjamini1995controlling},\cite{benjamini2001control}.}\label{tab:pvalues-ctrp_cv_pred}
	\vspace{5mm}
	\begin{tabular}{lcc}
		
					Methods & {\it cw}KBMF$_{single-view}$ & {\it cw}KBMF$_{multi-view}$ \\
					\hline
					{\it cw}KBMF$_{single-view}$ & -- & $3.662 \times 10^{-4}$ \\
					KBMF$_{multi-view}$ & -- & $1.464 \times 10^{-3}$ \\
					BMTMKL$_{multi-view}$ & -- & $1.708 \times 10^{-3}$ \\
					KPMF$_{single-view}$ & $3.662 \times 10^{-4}$ & $3.662 \times 10^{-4}$ \\
					MT-LR$_{single-view}$ & $7.324\times 10^{-4}$ & $3.662 \times 10^{-4}$ \\
					Baseline & $3.662 \times 10^{-4}$ & $3.662 \times 10^{-4}$ \\
					\hline
	\end{tabular}
\end{table}
\begin{figure*}[h]
	\centering
	\includegraphics[width=1\textwidth,height=1\textheight,keepaspectratio]{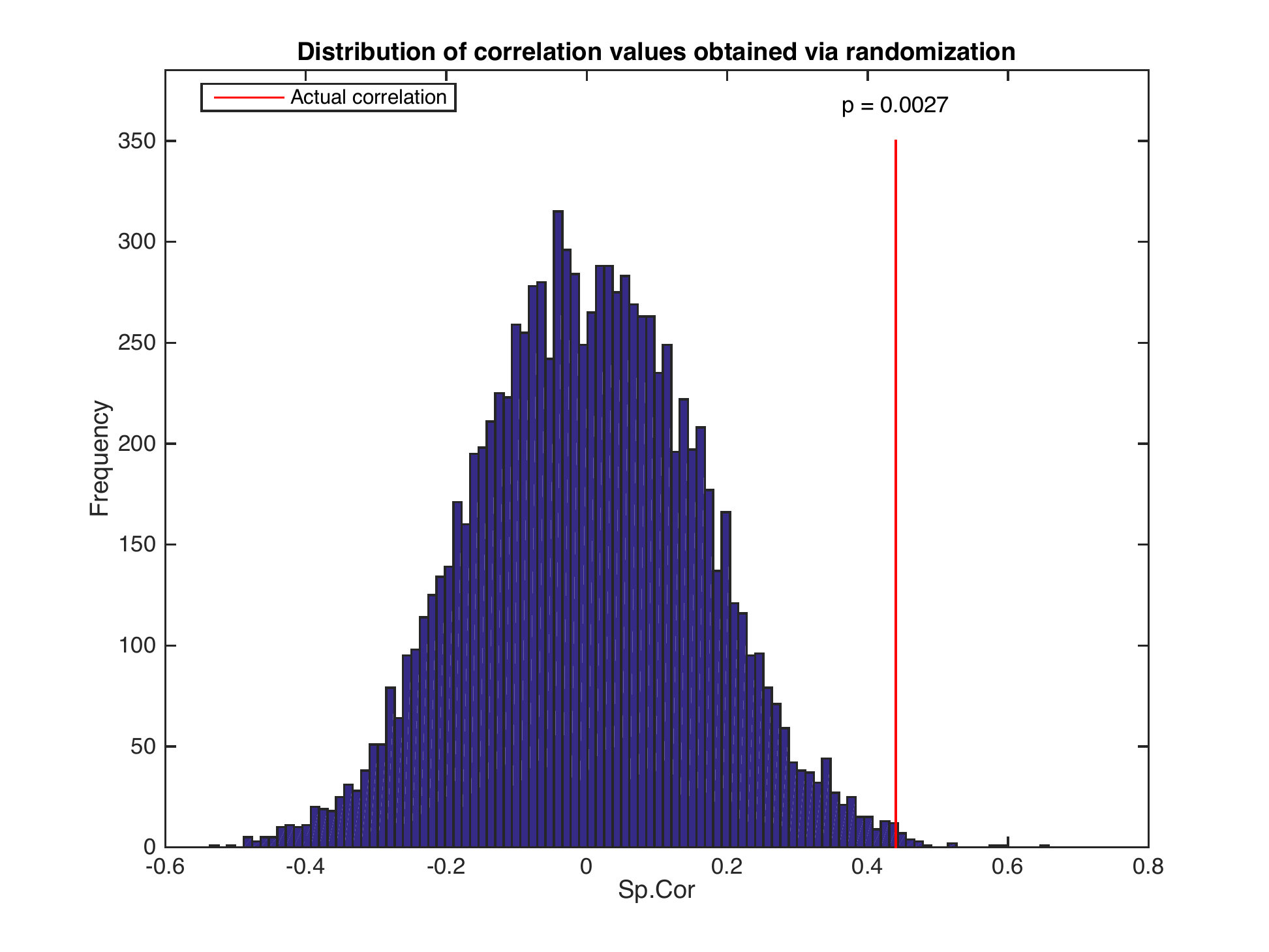}
	\caption{Statistical significance of {\it in silico} predictions obtained via randomizations test with 10000 random permutations of the drug responses. Spearman correlation $=$ 0.44 found to be significant with p $=$ 0.0027, than random distribution.}
	\label{fig:aml_validation_randomization}
\end{figure*}
\clearpage


	\end{document}